\documentclass[sigconf]{acmart}

\usepackage{xspace}
\usepackage{xcolor}
\usepackage{algpseudocode}
\usepackage{algorithm}
\usepackage{float}
\usepackage{amsmath}
\usepackage{algpseudocode}
\usepackage{makecell}
\usepackage{multirow}
\usepackage{tikz}
\usepackage{soul}
\usepackage{enumitem}  
\usetikzlibrary{shapes,arrows,positioning,calc,fit}
\usepackage{placeins}

\newcommand{\nop}[1]{}

\newcommand{\ouralg}{{\textsf{DualNILM}}\xspace}



\definecolor{Deepblue}{HTML}{1F77B4} 
\definecolor{Orange}{HTML}{FF7F0E} 
\definecolor{Green}{HTML}{2CA02C}    

\DeclareMathOperator*{\argmin}{arg\,min}

\usepackage{booktabs}
\usepackage{longtable}
\usepackage{tabularx}
\usepackage{fvextra}
\usepackage{array}
\DefineVerbatimEnvironment{ReviewBlock}{Verbatim}{%
  fontsize=\small,
  breaklines=true,
  breakanywhere=true,
  breaksymbolleft={}
}

\AtBeginDocument{%
  \providecommand\BibTeX{{%
    \normalfont B\kern-0.5em{\scshape i\kern-0.25em b}\kern-0.8em\TeX}}}

\setcopyright{acmlicensed}
\copyrightyear{2025}
\acmYear{2025}
\acmDOI{XXXXXXX.XXXXXXX}

\acmConference[Preprint, Accepted to e-Energy '26]{The Seventeenth ACM International Conference on Future and Sustainable Energy Systems}{June 23--25, 2026}{Banff, Canada}

\title[Energy Injection Identification enabled Disaggregation with Deep Multi-Task Learning]{Energy Injection Identification enabled Disaggregation with Deep Multi-Task Learning}

\author{Xudong Wang}
\orcid{0000-0001-7723-9966}
\affiliation{%
  \institution{The Chinese University of Hong Kong, Shenzhen}
  \country{}
}
\email{xudongwang@link.cuhk.edu.cn}

\author{Guoming Tang*}\thanks{*~Corresponding author.}
\orcid{0000-0001-9801-1055}
\affiliation{%
  \institution{The Hong Kong University of Science and Technology (Guangzhou)}
  \country{}
}
\email{guomingtang@hkust-gz.edu.cn}

\author{Junyu Xue}
\orcid{0009-0009-1119-326X}
\affiliation{%
  \institution{Southern University of Science and Technology}
  \country{}
}
\email{junyuxue@outlook.com}

\author{Srinivasan Keshav}
\orcid{0000-0002-6549-0464}
\affiliation{%
  \institution{University of Cambridge}
  \country{}
}
\email{sk818@cam.ac.uk}

\author{Tongxin Li}
\orcid{0000-0002-9806-8964}
\affiliation{%
  \institution{The Chinese University of Hong Kong, Shenzhen}
  \country{}
}
\email{litongxin@cuhk.edu.cn}

\author{Chris Ding}
\orcid{0009-0009-3374-1941}
\affiliation{%
  \institution{The Chinese University of Hong Kong, Shenzhen}
  \country{}
}
\email{chrisding@cuhk.edu.cn}

\renewcommand{\shortauthors}{X. Wang et al.}

\renewcommand\footnotetextcopyrightpermission[1]{}

\begin{document}

\begin{abstract}
Non-Intrusive Load Monitoring (NILM) offers a cost-effective method to obtain fine-grained appliance-level energy consumption in smart homes and building applications. However, the increasing adoption of behind-the-meter (BTM) energy sources such as solar panels and battery storage poses new challenges for conventional NILM methods that rely solely on at-the-meter data. The energy injected from the BTM sources can obscure the power signatures of individual appliances, leading to a significant decrease in NILM performance. 
To address this challenge, we present \textsf{DualNILM}, a deep multi-task learning framework designed for the dual tasks of appliance state recognition and injected energy identification. Using a Transformer-based architecture that integrates sequence-to-point and sequence-to-sequence strategies, \textsf{DualNILM} effectively captures multiscale temporal dependencies in the aggregate power consumption patterns, allowing for accurate appliance state recognition and energy injection identification. Extensive evaluation on self-collected and synthesized datasets demonstrates that \textsf{DualNILM} maintains an excellent performance for dual tasks in NILM, much outperforming conventional methods. Our work underscores the framework's potential for robust energy disaggregation in modern energy systems with renewable penetration. Synthetic photovoltaic augmented datasets with realistic injection simulation methodology are open-sourced at~\href{https://github.com/MathAdventurer/PV-Augmented-NILM-Datasets}{[\textit{link}]}.
\end{abstract}


\maketitle
\section{Introduction}\label{sec:intro}

Non-Intrusive Load Monitoring (NILM) technique enables disaggregation of total household energy consumption into individual appliance usage patterns~\cite{hart1992nonintrusive}, offering a cost-effective solution for fine-grained energy monitoring. This capability is increasingly critical for smart grid applications, facilitating targeted energy-saving recommendations, demand response programs, and enhanced grid efficiency~\cite{zoha2012non}.

Conventional NILM methods operate on a key assumption: the aggregate power measurements from the utility meter represent the sum of purely consumptive loads. Then these methods leverage unique electrical signatures of appliances to infer their operating states (or power) within the total load~\cite{makonin2015nonintrusive, batra2014nilmtk, rafiq2024review}.\nop{The value of accurate appliance-level monitoring becomes particularly pronounced during peak demand periods, where granular consumption insights enable utilities to implement precise demand response strategies and maintain grid stability without costly infrastructure upgrades~\cite{armel2013disaggregation}.}
However, the rapid adoption of behind-the-meter (BTM) renewable energy systems, particularly solar photovoltaics (PV), fundamentally disrupts this premise. The injection of energy from BTM sources creates bidirectional power flows that can dominate the aggregate power profile (especially during peak generation periods)~\cite{Saleem2022, rafiq2024review}, severely obscuring the signatures of individual appliances and challenging the foundation of traditional NILM.

\begin{figure}[t]
    \centering
    \includegraphics[width=1.00\linewidth, trim={0.2cm 0 0cm 0}, clip]{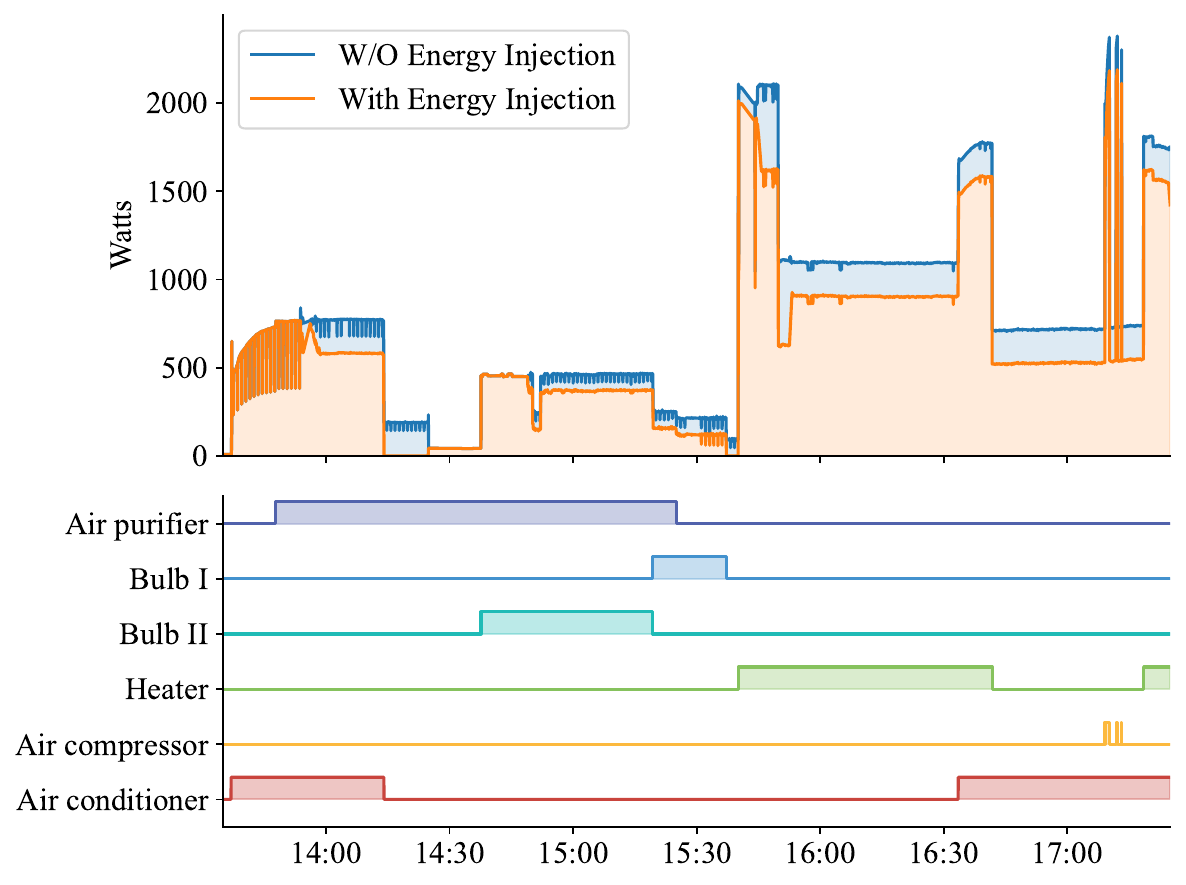}
    \caption{Laboratory measurements of aggregate power demonstrate the impact of controllable micro-inverter energy injection (0-500W adjustable output) on appliance signature visibility. The \textcolor{Deepblue}{blue} line represents baseline aggregate power without injection, \textcolor{Orange}{orange} shows the aggregate signal under controlled laboratory injection conditions. Bottom panels display ground-truth appliance ON/OFF states, illustrating how BTM energy injection can fundamentally obscure appliance state detection from the aggregate power signal.}
    \label{fig:1208aggregate_power}
    \vspace{-0.5cm}
\end{figure}

The resulting phenomenon of ``signal eclipse'' manifests itself through multiple interconnected challenges. \emph{At the signal level}, routine appliance operations become indistinguishable from generation fluctuations. For instance, a refrigerator's compressor activation, normally visible as a distinct step increase in aggregate power, can be entirely offset by a simultaneous dip in solar generation. \emph{At the temporal level}, natural solar variability, driven by cloud transients, atmospheric conditions and diurnal patterns, introduces non-stationary variations that mimic appliance switching events. This creates a perfect storm of ambiguity where traditional pattern recognition catastrophically fails. \emph{At the system level}, when generation exceeds consumption, the net power flow reverses. This changes the disaggregation problem from a geometrically constrained optimization within the non-negative orthant to a severely under-constrained inverse problem spanning the full real space.

The technical complexity of this challenge is significant. When PV generation exceeds instantaneous consumption, the disaggregation problem undergoes a fundamental mathematical transformation. Traditional NILM benefits from non-negativity constraints that bound the solution space. Each appliance contributes positively to the aggregate, creating a convex optimization landscape. In contrast, BTM generation shatters these constraints, expanding the feasible region to include arbitrary combinations of positive consumption and negative generation that yield identical net measurements. This transformation is further complicated by the electrical characteristics of modern grid-tied inverters, which operate at near-unity power factors to maximize generation efficiency.
As a result, BTM injection almost purely perturbs the active power\nop{~($S = P + jQ$ and the power factor is $\mathrm{PF} = P/|S| = \cos\phi$, operating at $\mathrm{PF} \approx 1$ means that the injection lies very close to the active power axis). Consequently, the active power trajectory is strongly distorted by injection}, whereas the measured reactive power is only weakly affected and remains largely governed by load behavior.\nop{ In practice, appliance signatures from active power become heavily obscured, while their reactive footprints are only slightly perturbed.} This asymmetry renders conventional active-only NILM approaches particularly fragile under BTM injection.

Figure~\ref{fig:1208aggregate_power} empirically demonstrates these theoretical challenges through controlled laboratory experiments. Our micro-inverter injection setup reveals how even modest energy generation fundamentally distorts the aggregate signal. During injection periods, the expected additive structure of load aggregation breaks down. The aggregate signal (orange) deviates dramatically from the consumption-only baseline (blue), while the actual appliance states (bottom panels) become completely obscured. This controlled experiment confirms that \emph{the challenge transcends simple noise addition, revealing a fundamental incompatibility between conventional NILM architectures and the bidirectional power flows of modern energy systems}.

From a utility and grid operator perspective, maintaining load visibility despite BTM resources is critical for demand response, grid stability, and planning. While smart inverters can theoretically report generation data, practical challenges include: ownership and privacy boundaries between customers and utilities, heterogeneous equipment with proprietary communication protocols, legacy installations lacking smart monitoring capabilities, and aggregated scenarios such as multi-tenant buildings or community solar arrangements without individual metering.

To address these challenges, we propose \textsf{DualNILM}, a deep multi-task learning framework specifically designed for the simultaneous disaggregation of appliance states and energy injection. Our key insight is that generation and consumption, rather than being antagonistic signals, contain \textit{complementary} information that can be leveraged for mutual disambiguation. The framework unifies sequence-to-point learning for discrete appliance state detection with sequence-to-sequence modeling for continuous injection estimation within a Transformer-based architecture. This design captures both short-term appliance switching events and long-term generation trends. A critical feature of \textsf{DualNILM} is its native processing of multi-channel feature sequences, incorporating active power, reactive power and other extensible signatures, to exploit cross-dimensional correlations and resolve ambiguities intractable in single-channel analysis. Furthermore, an attention mechanism dynamically weights these feature channels based on the instantaneous generation-consumption balance, effectively performing context-aware feature selection under dynamic conditions.

Our main contributions are:
\begin{itemize}[leftmargin=*, topsep=4pt, itemsep=2pt]
    \item \textbf{Systematic characterization of BTM challenges:} We analyze how behind-the-meter energy injection transforms the NILM problem from a constrained optimization to an under-determined inverse problem, formally identifying key failure modes such as signal eclipse and temporal aliasing.
    
    \item \textbf{Dual-task architecture design:} We propose \textsf{DualNILM}, a framework that jointly optimizes appliance recognition and energy injection estimation. Its attention-based multi-channel feature fusion mechanism dynamically leverage complementary information across electrical signatures, enabling robust disaggregation under eclipsed signals.
    
    \item \textbf{Comprehensive empirical validation:} We demonstrate \textsf{DualNILM}'s state-of-the-art performance across diverse operational regimes~with heterogeneous BTM injection sources, from controlled laboratory experiments (500W micro-inverter) to synthetic residential-scale scenarios (2kW PV systems).
    
    \item \textbf{Open research infrastructure:} We release the synthetic PV-augmented versions (with realistic weather data) of REDD and UK-DALE datasets and the corresponding benchmark suites to support future research in this critical area~\href{https://github.com/MathAdventurer/PV-Augmented-NILM-Datasets}{[\textit{link}]}\footnote{\url{https://github.com/MathAdventurer/PV-Augmented-NILM-Datasets}}.
\end{itemize}

\nop{
The significance of this work extends beyond addressing an immediate technical challenge. With global residential solar capacity projected to exceed 5.5 TW by 2030, representing a threefold increase from 2023 levels~\cite{IEA2023solar, IRENA2023outlook}, the ability to maintain load visibility despite bidirectional power flows becomes critical infrastructure for grid intelligence. The dual-task paradigm we propose provides a scalable framework naturally extensible to emerging energy scenarios involving battery storage, vehicle-to-grid systems, and peer-to-peer energy trading. By demonstrating that generation and consumption can be simultaneously tracked using existing smart meter infrastructure, without requiring expensive smart inverters or dedicated generation meters, this work democratizes access to advanced energy analytics and accelerates the transition toward a truly intelligent and sustainable energy future.
}

\section{Related work}\label{sec:related_work}

\paragraph{\textbf{Evolution of NILM Technique}}
Non-Intrusive Load Monitoring (NILM) has been a subject of extensive research since its introduction by Hart~\cite{hart1992nonintrusive}. Traditional NILM approaches rely on identifying unique appliance signatures in the aggregate power signal to disaggregate individual appliance usage~\cite{makonin2015nonintrusive}. Techniques such as Hidden Markov Models (HMMs)~\cite{kim2011unsupervised}, factorial HMMs~\cite{kolter2012approximate}, and combinatorial optimization~\cite{kolter2011redd} have been widely used. With the advent of deep learning, neural network-based methods have significantly advanced NILM performance. Convolutional Neural Networks (CNNs) and Recurrent Neural Networks (RNNs) have been employed to capture temporal dependencies and extract features from power signals~\cite{kelly2015neural, zhang2018sequence, ayub2023contextual}. Sequence-to-point and sequence-to-sequence models have been proposed to improve disaggregation accuracy~\cite{zhang2018sequence, du2016regression}. Recently, researchers have also introduced advanced deep architectures, such as transformer-based models and graph neural networks, to further enhance disaggregation performance~\cite{ccavdar2021efficient, shang2024graphnilm}. These approaches leverage long-range temporal context and inter-appliance relationships to achieve higher accuracy on benchmark datasets while improving the generalization ability of NILM models.

\paragraph{\textbf{Categories of NILM Methods}}
NILM methods can also be categorized according to their task-oriented objectives. Many approaches focus on \textbf{event-based detection}, where the key aim is to recognize appliance switching events (ON/OFF transitions) and label those events with the corresponding device~\cite{wang2022evsense, liaqat2022emeda}. This category typically relies on detecting changes in aggregate power that exceed a certain threshold, followed by a classification stage to identify which appliance caused the event. Various improvements to event detection have been proposed, including hybrid schemes that combine primary event triggers with secondary verification algorithms to reduce false detections~\cite{lu2019hybrid}. By isolating transition points in time series, event-based NILM methods can be highly interpretable and computationally efficient, but may struggle with appliances that exhibit gradual power variations or multiple internal states~\cite{meziane2017event}. On the other hand, the \textbf{steady-state-based approaches} emphasize continuous monitoring of power consumption, estimating the exact load of each device over time. These NILM methods often employ more complex models (e.g., HMMs or deep neural networks) and can capture multi-level operating states of appliances~\cite{cao2023multitask}.

\paragraph{\textbf{NILM Methods amid RE or BTM Integration}}

Previous work addressing NILM with the integration of renewable energy is limited. In~\cite{hamidi2016power}, the authors investigate methods for managing power injection systems in renewable setups, emphasizing the stability of the system. Similarly, Aygen \textit{et al.}~\cite{aygen2020zero} consider a zero-sequence current injection technique to mitigate unbalanced loads in systems integrated with renewables, which, however, is not directly applicable to NILM. Only a few studies have attempted to explicitly disaggregate solar generation from the net metered signal. SunDance~\cite{chen2017sundance} from Chen \& Irwin presents a black-box approach specifically for solar disaggregation, combining physical solar generation models with machine learning to map weather conditions to solar output without appliance-level monitoring. Noori \textit{et al.}~\cite{ra2023behind} recently examined solar disaggregation and highlighted that incorporating an external context, such as solar irradiance, can improve the detection of solar generation. In~\cite{2020nilmsolardisaggregation}, the authors investigate solar energy disaggregation but rely on high-frequency data and additional features, which may not be practical in real-world deployments. Li \textit{et al.}~\cite{li2024bert} propose BERT-based architectures for behind-the-meter PV monitoring, utilizing user embeddings and weather features for power regression tasks. Other studies have considered the use of additional sensors or prior knowledge of the energy injection~\cite{abir2021iot}, which limits their applicability.

\nop{Multi-task learning has been explored in NILM to predict multiple appliance states simultaneously~\cite{kaselimi2019multi}. SAMNet~\cite{liu2022samnet} employs an attention-based multi-task framework to enable low-latency disaggregation. UNetNILM~\cite{faustine2020unet} introduced a sophisticated approach to multi-task learning in NILM, utilizing a U-Net architecture to detect appliance states and estimate their power consumption. Recently, Gkalinikis \textit{et al.}~\cite{virtsionis2023variational} proposed a variational multi-target regression model to jointly estimate multiple appliance power signals, further demonstrating the benefits of holistic disaggregation models. However, these methods typically do not consider the presence of energy injection and focus on appliances with positive power consumption.} 

Our work differs from existing approaches by explicitly addressing energy injection challenges through deep multi-task learning. \textsf{DualNILM} jointly performs appliance state recognition and injection disaggregation without requiring additional sensors or high-frequency data. Critically, given the scarcity of NILM datasets containing renewable generation, we also provide synthetic augmentation methodology/datasets that incorporate realistic PV injection patterns derived from actual meteorological data, enabling reproducible research in this emerging domain.

\section{Problem Definition}
\label{sec:problemformulation}


Traditional NILM approaches infer appliance-level power consumption from aggregated household measurements under the assumption that all loads are strictly non-negative~\cite{hart1992nonintrusive, kolter2011redd}. However, modern residences increasingly incorporate BTM renewable generation, where micro-inverters inject power back to the grid, fundamentally violating this core assumption. 
\nop{Our laboratory studies demonstrate that even modest 500W micro-inverter injection significantly distorts aggregate load profiles, rendering appliance signatures indistinguishable (Figure~\ref{fig:1208aggregate_power}) and causing substantial disaggregation errors for conventional methods. Moreover, the difficulty can be exacerbated by low sampling rates or the lack of multi-feature data, both common in real-world metering systems~\cite{kelly2015neural, rafiq2024review}.}

\subsection{NILM with BTM Energy Injection}

In classical NILM, aggregate power $y(t)$ at time $t$ for $N$ appliances follows:
\begin{equation}
y(t) = \sum_{n=1}^N s_n(t) \cdot x_n(t) + \epsilon(t),
\label{eq:classic-nilm}
\end{equation}
where $s_n(t)\in\{0,1\}$ denotes appliance state, $x_n(t)\geq 0$ represents power consumption when active, and $\epsilon(t)$ is measurement noise. Since NILM typically monitors a subset of $K$ target appliances, this becomes:
\begin{equation}
y(t) = \sum_{n=1}^{K} s_n(t) \cdot x_n(t) + v(t),
\label{eq:classic-subset}
\end{equation}
where $v(t)$ aggregates remaining loads and noise. While solving for $\{s_n(t), x_n(t)\}$ is NP-hard~\cite{hart1992nonintrusive}, the non-negative structure historically enables acceptable performance through heuristic and learning-based methods~\cite{makonin2015nonintrusive, batra2014nilmtk}.

With the BTM injection, the aggregate signal transforms to:
\begin{equation}
y(t) = \sum_{n=1}^{K-1} s_n(t) \cdot x_n(t) - \pi(t) + v(t),
\label{eq:with-injection}
\end{equation}
where $\pi(t) \geq 0$ represents energy injection. This modification fundamentally alters the problem structure: the feasible solution space expands beyond the non-negative orthant, breaking geometric constraints that classical factorization methods rely upon. The resulting identifiability problem manifests itself as increased ambiguity in explaining power dips, degraded solution stability, and insufficient discriminative information in single-channel measurements. (See Appendix~\ref{appendix:challenge_analysis} for detailed theoretical analysis.)

\subsection{DNN-based NILM with Dual Tasks}

Deep neural networks (DNNs) have shown promise in NILM by exploiting additional features or electrical signatures beyond active power~\cite{kelly2015neural, zhang2018sequence}. Let $\mathbf{f}(t)\in\mathbb{R}^{F}$ denote the multi-dimensional feature vector at time $t$, including active power, reactive power, current harmonics, etc. A neural model processes these features as:
\begin{equation}
[\hat{s}_1, \ldots, \hat{s}_K, \hat{x}_1, \ldots, \hat{x}_K] = \mathrm{DNN}_\theta(\mathbf{y}, \mathbf{f}),
\label{eq:dnn-general}
\end{equation}
where $\theta$ represents trainable parameters. These richer feature spaces provide crucial discriminative information, as appliances and injection sources exhibit distinct signatures across multiple electrical dimensions. However, directly applying standard architectures to the injection-augmented problem remains challenging due to the different nature of consumption versus generation patterns.


Recognizing that energy injection exhibits distinct temporal and electrical characteristics from appliance consumption, we propose treating injection estimation as a dedicated task within a multi-task learning target: 
\begin{equation}
\Bigl[\hat{s}_1, \ldots, \hat{s}_K,\; \hat{\pi}(t)\Bigr]
\;=\;
\mathrm{DNN}_\theta\!\bigl(\mathbf{y},\,\mathbf{f}\bigr),
\end{equation}
in which the following dual tasks are simultaneously tackled:

\begin{itemize}
    \item \textbf{Appliance State Recognition:} Identify ON/OFF states for target appliances $\hat{s}_1, \ldots, \hat{s}_K$, using learned representations that capture appliance-specific switching patterns and steady-state signatures.
    \item \textbf{Energy Injection Disaggregation:} Estimate injection power $\hat{\pi}(t)$, through a specialized pathway that accounts for their continuous nature and reduces the solution search space to pursue a stable optimization landscape.
\end{itemize}

This formulation enables the model to learn complementary representations: discrete state transitions for appliances and continuous power variations for injection. By explicitly modeling injection rather than treating it as negative consumption, the framework naturally accommodates bidirectional power flows while reducing optimization complexity. The dual-task NILM formulation is also well-suited for DNN-based NILM models that can leverage rich features, capture complex patterns and dependencies in the data, allowing accurate estimation of the appliance states and the energy injection~\cite{kelly2015neural, zhang2018sequence, kaselimi2019multi}.

The following section presents the architecture of our approach that implements this dual-task paradigm via specialized neural pathways, each optimized for its respective disaggregation challenge.

\begin{figure*}[!t]
    \centering
    \fbox{
    \includegraphics[width=1.0\linewidth]{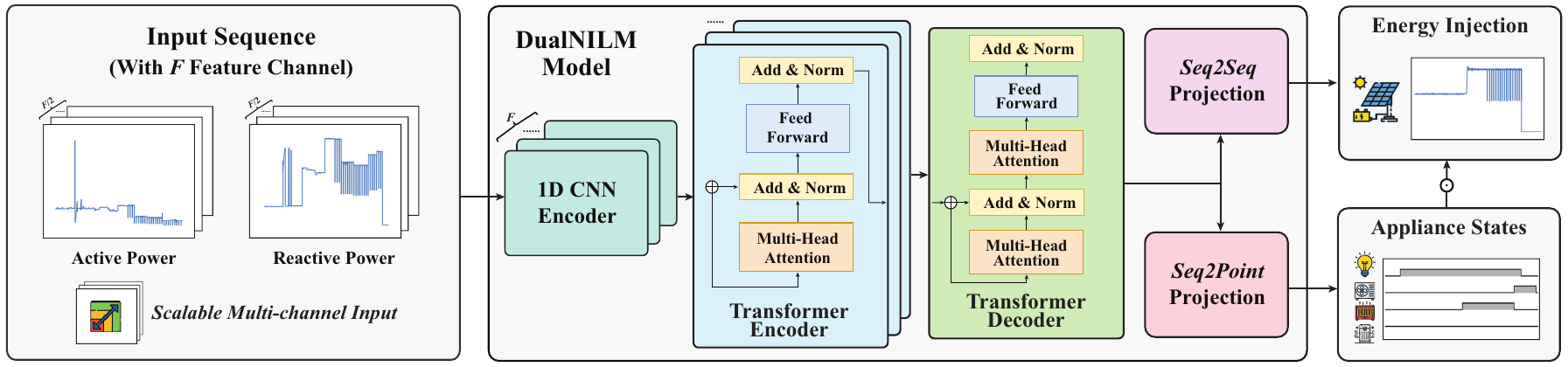}}
    \caption{Illustration of our proposed \ouralg.}
    \label{fig:DualNILM-architecture}
\end{figure*}

\section{\ouralg}
\label{sec:proposedmethod}

To address the challenges described, we propose \ouralg, a novel deep multi-task learning framework for simultaneous energy disaggregation and injection identification. This section details the architecture of \ouralg, its key components, and the motivations behind our design choices. More implementation details can be found in the Appendix~\ref{appendix:modeldesign}. 

\subsection{Design Rationale}
\label{subsec:method-overview}

The behind-the-meter negative power flows from micro-inverters expand the solution space beyond the typical nonnegative constraint, compounding disaggregation ambiguity. \ouralg\ follows the Dual-Task NILM manner and mitigates this difficulty by:
\begin{itemize}
    \item \textbf{Separately Modeling Injection:} Instead of forcing a single model head to predict all appliance consumptions (including negative flows), we assign a distinct output channel for the injection source. This explicit handling of negative values helps isolate the behind-the-meter signal from conventional appliances.
    \item \textbf{Multiscale Feature Extraction:} Appliance usage patterns often manifest as short-term events (e.g., ON/OFF transients) superimposed on longer-term trends, whereas injection can exhibit relatively smooth variations (e.g., daily solar generation). To capture both fine-grained patterns and global context, \ouralg~combines convolutional layers (for local feature learning) with Transformer layers (for longer-range sequence dependencies).
    \item \textbf{Multi-Task Optimization:} The model jointly learns to classify appliance states (binary ON/OFF) and predict the magnitude of the injected power. This multi-task approach leverages shared feature representations, improved regularization, and explicit constraints (such as maximum injection capacity) to handle overlapping load signatures more effectively.
\end{itemize}

\subsection{Model Architecture}

The overall architecture of \ouralg is illustrated in Figure~\ref{fig:DualNILM-architecture}. Our model comprises four main components: 1) CNN encoders, 2) transformer encoders, 3) a transformer decoder, and 4) task-specific projection layers. This design allows for the effective capture of both local and global temporal dependencies in the power consumption patterns while enabling multi-task learning.

\paragraph{\textbf{CNN Encoders.}}
The input to our model is a multi-channel time series $\mathbf{X} \in \mathbb{R}^{T \times F}$, where $T$ is the sequence length and $F$ is the number of input features (e.g., active power, reactive power). We employ a separate 1D CNN Encoder for each feature channel to extract local temporal patterns. Each CNN Encoder consists of multiple 1D convolutional layers interleaved with LayerNorm operations:
\begin{equation}
\mathbf{H}_f = \text{CNN}_f(\mathbf{X}_f), \quad f = 1, \ldots, F
\end{equation}
where $\mathbf{X}_f \in \mathbb{R}^T$ is the $f$-th feature channel and $\mathbf{H}_f \in \mathbb{R}^{T \times D}$ is the corresponding encoded representation. Using separate CNN Encoders for each feature allows the model to learn feature-specific local patterns, enhancing its ability to capture the unique characteristics of different electrical measurements \cite{kelly2015neural}.

\paragraph{\textbf{Transformer Encoder.}}
The outputs of the CNN Encoders are concatenated along the feature dimension and fed into Transformer Encoders \cite{vaswani2017attention}. The Transformer Encoder employs multi-head attention mechanisms to capture long-range dependencies in the input sequence:
\begin{equation}
\mathbf{Z} = \text{TransformerEncoder}([\mathbf{H}_1; \ldots; \mathbf{H}_F])
\end{equation}
where $\mathbf{Z} \in \mathbb{R}^{T \times D'}$ is the output of the Transformer Encoder. This component allows \textsf{DualNILM} to model complex interactions between different appliances and the energy injection source over time, which is crucial for accurate disaggregation and injection estimation \cite{zhang2018sequence}.

\paragraph{\textbf{Transformer Decoder \& Task-Specific Projections.}}
To address the dual tasks of appliance state detection and energy injection estimation, we employ a hybrid approach combining sequence-to-point (Seq2Point) and sequence-to-sequence (Seq2Seq) strategies:
\begin{itemize}
    \item 
    For appliance state detection, we use a Seq2Point approach. The last time step of the Transformer Encoder output is projected to predict the states of certain appliances $i$:
    \begin{equation}
    \hat{\mathbf{s_i}} = \sigma(\mathbf{W}_{s_i} \mathbf{z}_T + \mathbf{b}_{s_i})
    \end{equation}
    where $\mathbf{z}_T$ is the last time step of $\mathbf{Z}$, $\mathbf{W}_{s_i}$ and $\mathbf{b}_{s_i}$ are learnable parameters, and $\sigma$ is the sigmoid activation function.
    \item 
    For energy injection estimation, we employ a Seq2Seq approach using a transformer decoder:
    \begin{equation}
    \hat{\mathbf{x}}_K = \sigma(\mathbf{W}_x \cdot \text{TransformerDecoder}(\mathbf{Z}) + \mathbf{b}_x)
    \end{equation}
    where $\hat{\mathbf{x}}_K \in \mathbb{R}^T$ is the predicted energy injection sequence, and $\mathbf{W}_x$ and $\mathbf{b}_x$ are learnable parameters. 
    \item
    For enhanced cross-task synergy, the final injection estimate incorporates the predicted inverter state through element-wise filtering:
\begin{equation}
\hat{\mathbf{x}}_K^{\text{final}} = \hat{\mathbf{x}}_K \odot \hat{\mathbf{s}}_K
\end{equation}
where $\odot$ denotes the Hadamard product. This cross-task constraint ensures injection predictions are zeroed when the inverter is predicted inactive, fundamentally improving the physical consistency of the dual-task outputs.
\end{itemize}

\paragraph{\textbf{Design Justification for \ouralg}}
\label{subsec:design-justification}

The proposed architecture addresses the signal eclipse phenomenon directly through the above coupled design choices, ensuring both theoretical soundness and practical viability. \underline{\emph{First}}, the CNN-Transformer hybrid accommodates distinct temporal characteristics: appliance switching events exhibit sharp transients captured by convolutional kernels, while energy injection exhibits complex multiscale temporal dependencies and cross-correlation patterns requiring long-range attention mechanisms to capture their dynamics.
\underline{\emph{Second}}, explicit separation of injection estimation through dedicated Seq2Seq projection prevents mathematical contamination of appliance state classification, where negative flows would corrupt binary decision boundaries. This architectural separation, combined with LayerNorm stabilization~\cite{ba2016layernormalization}, enables direct processing of raw meter readings while maintaining modular scalability for incremental appliance addition.
\underline{\emph{Third}}, the shared encoder representations enable cross-task synergy that significantly reduces the problem of identification, allowing accurate disambiguation between actual load reduction and injection-induced power dips within their respective constraint domains.

\subsection{Training Objective and Optimization}
Denote $\mathbf{Y}$ is the set of ground truth, e.g $(\mathbf{S},\mathbf{X})$. Our optimization target is: 
\begin{equation}
\argmin_{\theta} \mathcal{L}(\mathbf{Y}, \text{\ouralg}_{\theta}([\mathbf{y}, \mathbf{f}]))
\end{equation}
The multi-task nature of \ouralg is reflected in its loss function~$\mathcal{L}$, which combines binary classification/recognition loss for appliance state detection and the disaggregation error loss for energy injection estimation:
\begin{equation}
\mathcal{L} =  \mathcal{L}_{\text{Rec}}(\hat{\mathbf{S}}, \mathbf{S}) + \lambda \cdot \mathcal{L}_{\text{Disagg}}(\hat{\mathbf{x}}_K, \mathbf{x}_K)
\end{equation}
where $\lambda\in (0,1]$ is the weighting factor to balance the dual targets. 

Our model can be optimized using gradient backpropagation approaches such as SGD~\cite{amari1993backpropagation}, Adam~\cite{kingma2014adam}, etc. Specifically, we employ Dice loss~\cite{milletari2016v}, which is particularly effective for imbalanced binary classification problems for appliance state recognition:
\begin{equation}
\mathcal{L}_{\text{Rec}} = \sum_{i = 1}^{K} (1 - \frac{2|\hat{\mathbf{s_i}} \cap \mathbf{s_i}|}{|\hat{\mathbf{s_i}}| + |\mathbf{s_i}|})
\end{equation}

For injection estimation loss, we use the standard L2 measure: Root Mean Square Error (RMSE). Note that we can apply constraints/normalization on injection disaggregation targets as the maximum output of micro-inverters or community solar installations typically must adhere to standardized values like~EN50583~\cite{EN50583}.

\section{Experimental Setup}\label{sec:experimentalsetup}

To thoroughly evaluate the performance of \ouralg, we conducted extensive experiments using both real-world data collected in our laboratory and simulated data based on widely used public datasets. This section details our experimental setup, including datasets and simulations, evaluation metrics, benchmark methods, and experiment procedures. 

\subsection{Datasets and Simulation}\label{sec:datasetandsim}

\paragraph{\textbf{Laboratory Dataset}}

We collected a real-world dataset in our laboratory and conducted extensive experiments. The dataset includes individual power consumption data for multiple appliances and the BTM energy injection from a micro-inverter. The data was collected from December 6 to December 15, 2023, and was sampled at 2-second intervals using sub-meters for each appliance on both active power and reactive power. The dataset includes the following appliances: i) Air purifier, ii) Heater, iii) Light bulb Type I, iv) Light bulb Type II, v) Air compressor, vi) Air conditioner, vii) Micro-inverter (energy injection).
The laboratory data collection employed a combination of automated random control and manual random triggering to eliminate temporal correlations. For each of the included appliances, operating states were randomly selected, along with randomized run durations and random micro-inverter power level adjustments. This creates an \emph{adversarial scenario testing pure disaggregation capability without exploitable temporal patterns}. Table~\ref{tab:dataset_stats} presents the key statistics of our laboratory dataset. 

\begin{table}[h]
\centering
\caption{Statistics of the Laboratory Dataset}
\label{tab:dataset_stats}
\resizebox{\linewidth}{!}{
\setlength{\tabcolsep}{1.2mm}
\renewcommand{\arraystretch}{0.97}
\begin{tabular}{lccc}
\hline
\textbf{Appliance} & \textbf{Mean Power (W)} & \textbf{Std Dev (W)} & \textbf{On-time (\%)} \\
\hline
Air purifier & 34.25 & 2.35 & 40.57 \\
Heater & 1345.73 & 476.90 & 28.69\\
Light bulb Type I & 204.84 & 13.64 & 10.54 \\
Light bulb Type II & 412.99 & 49.39 & 10.59 \\
Air compressor & 1408.04 & 287.72 & 0.4885 \\
Air conditioner & 695.59 & 82.92 & 38.61 \\
Micro-inverter & 219.31 & 115.77 & 63.56 \\
\hline
\multicolumn{4}{l}{\textit{\footnotesize{Micro-inverter's value means the power injection}}}
\end{tabular}}
\end{table}

Furthermore, in our experiments on our laboratory datasets, we perform cross-validation by using data from one day for testing and the data from the remaining days for training, iterating over all days in the dataset.

\begin{figure}[t]
    \centering
    \includegraphics[width=1\linewidth]{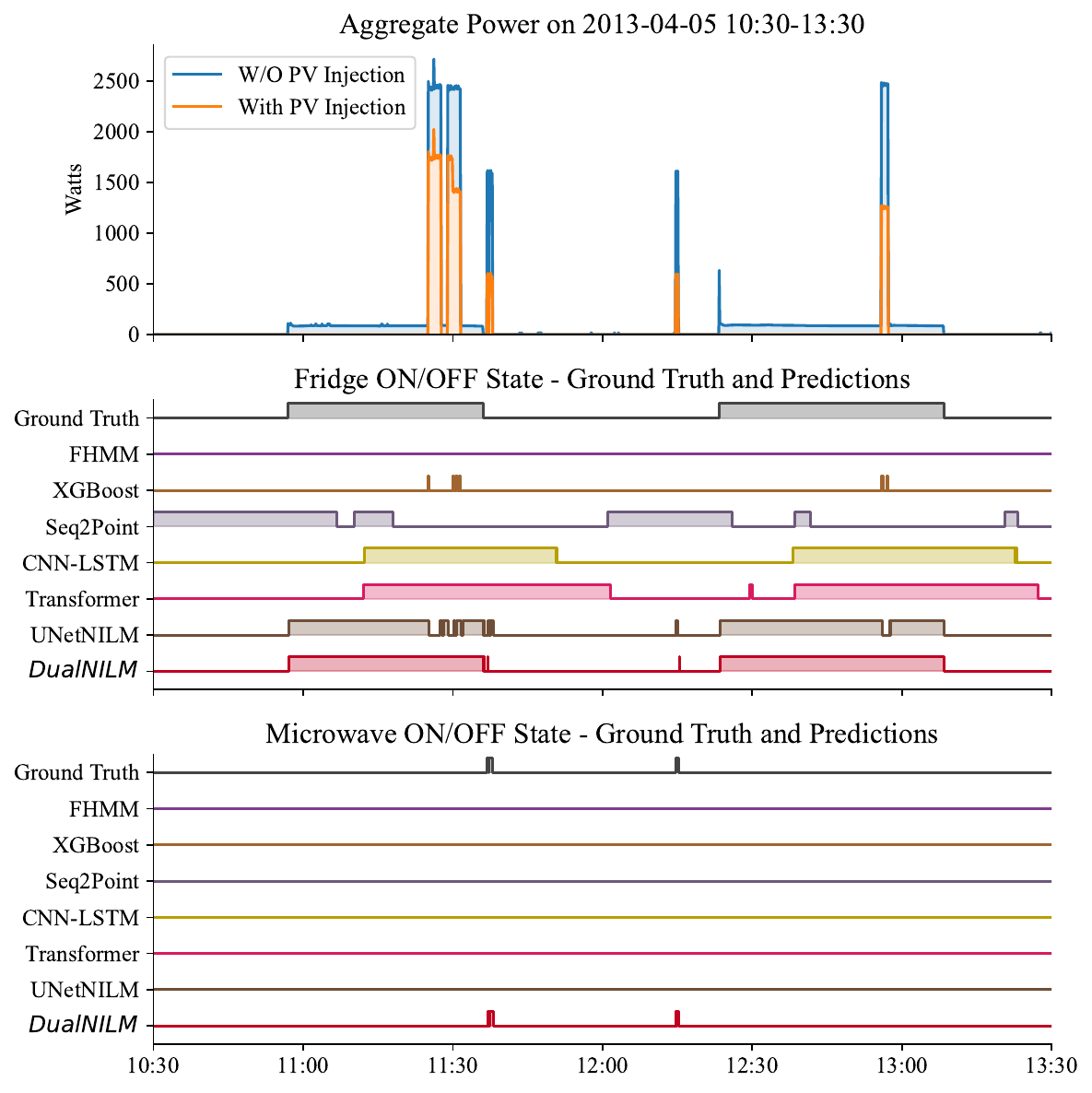}
    \caption{Simulated Photovoltaic energy injection on UKDALE House 1 on Apr.~05, 2013, and the corresponding Fridge and Microwave's ON/OFF states and their predictions from \ouralg and benchmarks.}
    \label{fig:ukdale1_simulation_demo}
\end{figure}

\paragraph{\textbf{Synthetic Public Datasets}}\label{subsec:simulated_injection}

To evaluate the generalizability of \ouralg{} and facilitate comparison with existing methods, we simulate energy injection scenarios using two widely-used public NILM datasets: REDD~\cite{kolter2011redd} and UK-DALE~\cite{kelly2015uk}. Our simulation takes advantage of real-world solar irradiance data and photovoltaic (PV) system modeling to create realistic scenarios of residential solar energy injection. Detailed implementation is open-source and available in ~\href{https://github.com/MathAdventurer/PV-Augmented-NILM-Datasets}{[\textit{link}]}. The simulation process involved the following steps:

\begin{enumerate}
\item \textbf{Selection of Houses:} Select houses from each dataset with suitable appliance combinations. Specifically, we focused on houses with the appliances commonly used in prior works:
microwave, fridge, dishwasher, and washing machine for REDD; and kettle, microwave, fridge, dishwasher, and washing machine for UK-DALE.

\item \textbf{Acquisition of Solar Irradiance Data:} We obtain historical solar irradiance data from the National Solar Radiation Database (NSRDB) provided by the National Renewable Energy Laboratory (NREL)~\cite{nsrdb2020}. The NSRDB offers detailed solar radiation data, including Global Horizontal Irradiance (GHI), Direct Normal Irradiance (DNI), and Diffuse Horizontal Irradiance (DHI), with a temporal resolution of 30 minutes. We collect data for the rough geographical locations corresponding to the datasets:

\begin{itemize}
    \item \textbf{REDD Location:} Boston, MA, USA (Latitude: 42.3601$^\circ$N, Longitude: 71.0589$^\circ$W)
    \item \textbf{UK-DALE Location:} London, UK (Latitude: 51.5074$^\circ$N, Longitude: 0.1278$^\circ$W)
\end{itemize}

\item \textbf{Simulation of Photovoltaic System Output:} Using the collected irradiance data, we simulate a residential PV system with a designed maximum output capacity of 120~W, representing a small-scale household solar installation. The simulation accounts for the effects of temperature on the efficiency of PV, following the methods described in~\cite{duffie2013solar, marion2002pvwats}.

The cell temperature $T_{\text{cell}}$ is calculated as:
\begin{equation}
T_{\text{cell}} = T_{\text{ambient}} + \frac{\text{GHI}}{1000} \times \left( \frac{T_{\text{NOCT}} - 20}{0.8} \right),
\end{equation}
where $T_{\text{ambient}}$ is the ambient temperature in degrees Celsius, GHI is the Global Horizontal Irradiance in W/m$^2$, and $T_{\text{NOCT}}$ is the Nominal Operating Cell Temperature, typically 45$^\circ$C. The efficiency adjustment $\eta_{\text{adj}}$ due to temperature is computed as:
\begin{equation}
\eta_{\text{adj}} = 1 + \gamma \times (T_{\text{cell}} - 25),
\end{equation}
where $\gamma$ is the temperature coefficient ($-0.005$~$^\circ$C$^{-1}$ for silicon-based PV modules). The PV power output $P_{\text{PV}}$ is then calculated using:
\begin{equation}
P_{\text{PV}} = \min\left( P_{\text{rated}}, \frac{\text{GHI} \times P_{\text{rated}}}{1000} \times \eta_{\text{adj}} \times \eta_{\text{inv}} \right),
\end{equation}
where $P_{\text{rated}}$ is the rated capacity (120~W in our simulation) and $\eta_{\text{inv}}$ is the inverter efficiency (0.96 based on~\cite{IEC60034-1:2022}).

\item \textbf{Reactive Power Simulation:} For comprehensive power analysis, we simulate reactive power characteristics for both appliances and the PV system based on typical power factors from industry standards~\cite{IEEE1459-2010}. For each appliance with active power $P$, its reactive power $Q$ is calculated as:

\begin{equation}
Q = P \times \tan(\arccos(\text{PF}))
\end{equation}

where PF varies by appliance type following established standards~\cite{Masoum2015, Singh2003}: refrigerator (0.85), microwave (0.95), washing machine (0.75), dishwasher (0.80), kettle (0.99), and micro-inverter (0.98).

\item \textbf{Integration with Household Consumption:} We integrate the simulated PV output with the aggregate power consumption data from the datasets. To reflect realistic scenarios where excess generation is not exported to the grid, we cap the PV output at the household's instantaneous consumption:
\begin{equation}
P_{\text{PV}}^{\text{adjusted}} = \min\left( P_{\text{PV}}, P_{\text{agg}} \right),
\end{equation}
where $P_{\text{agg}}$ is the original aggregate power consumption.
The aggregate power with PV injection is then:
\begin{equation}
P_{\text{agg}}^{\text{injected}} = P_{\text{agg}} - P_{\text{PV}}^{\text{adjusted}}.
\end{equation}
We ensure that $P_{\text{agg}}^{\text{injected}} \geq 0$ by setting any negative values to zero.
\end{enumerate}
Figure~\ref{fig:ukdale1_simulation_demo} shows an example to demonstrate a challenging disaggregation scenario on UKDALE House 1 during the midday hours (10:30-13:30) on April 05, 2013, when PV injection interacts with typical household consumption. The aggregate power exhibits dramatic variations as appliances cycle, yet the PV injection (green shaded area) creates subtle but critical distortions in the net measurement. During periods of low baseline consumption, such as when only the refrigerator operates between switching events, even modest PV generation can mask appliance signatures entirely. 

We use the following cross-validation setup for each dataset:
\begin{itemize}
    \item REDD: Houses 1, 2, and 3: 1-week training, 3-day testing.
    \item UK-DALE: Houses 1 and 2: 2-week training, 1-week testing.
\end{itemize}

Table~\ref {tab:sim_dataset_stats_2kw} in the Appendix shows the key statistics of our synthetic datasets. More detailed dataset descriptions on both laboratory and simulated public data, including detailed configuration and simulation and appliance state labeling, are described in Appendix~\ref{appendix:dataset}.

\subsection{Evaluation Metrics}\label{sec:evalmetric}
\paragraph{\textbf{Appliance State Recognition Metrics}} For appliance state recognition tasks, we report Accuracy, Recall, Precision, and F1-score. Since the imbalanced nature of the appliance ON/OFF state, the F1-score serves as the most comprehensive metric.

All above metrics are calculated for each appliance individually and then averaged across all appliances to provide an overall performance assessment.

\paragraph{\textbf{Energy Injection Estimation Metric}}

For the BTM energy injection estimation task, we use the Root Mean Square Error (RMSE)  and Mean Absolute Error (MAE) as the metrics:

\begin{equation}
	\begin{aligned}
    RMSE  = \sqrt{\frac{1}{T}\sum_{t=1}^{T}(x_K(t) - \hat{x}_K(t))^2},\quad
     MAE  = \frac{1}{T}\sum_{t=1}^{T}|x_K(t) - \hat{x}_K(t)|
     \end{aligned}
\end{equation}
where $x_K(t)$ is the true injection value and $\hat{x}_K(t)$ is the estimated injection value at time $t$. These metrics quantify the accuracy of the model's energy injection estimates over time.

\subsection{Benchmark Methods}

To comprehensively evaluate \textsf{DualNILM}, we compare it with representative and state-of-the-art methods in both appliance state recognition and energy injection disaggregation tasks. Table~\ref{tab:benchmark_methods} summarizes our benchmark methods. 

\begin{table*}[t]
\centering
\caption{Summary of Benchmark Methods}
\label{tab:benchmark_methods}
\vspace{-5pt}
\resizebox{0.73\textwidth}{!}{
\setlength{\tabcolsep}{1.2mm}
\renewcommand{\arraystretch}{0.97}
\begin{tabular}{lllll}
\toprule
\textbf{Category} & \textbf{Method} & \textbf{Key Characteristics} & \textbf{Primary Strengths} & \textbf{Tasks}* \\
\midrule
\multirow{3}{*}{Traditional} 
& FHMM~\cite{kim2011unsupervised} & Parallel hidden Markov chains & Interpretable state modeling & Both \\
& XGBoost~\cite{chen2016xgboost} & Tree-based ensemble & Non-linear pattern recognition & State Recognition \\
& SunDance~\cite{chen2017sundance} & Physical solar modeling & Black-box estimation & Disaggregation \\
\midrule
\multirow{7}{*}{\makecell[l]{Deep\\Learning}} 
& Seq2Point~\cite{zhang2018sequence} & CNN-based architecture & Local temporal features & State Recognition \\
& CNN-LSTM~\cite{kaselimi2020multi} & Hybrid CNN-LSTM & Local-global temporal modeling & State Recognition \\
& Transformer~\cite{vaswani2017attention} & Self-attention mechanism & Long-range dependencies & State Recognition \\
& Seq2Seq~\cite{du2016regression} & Encoder-decoder LSTM & Sequence mapping & Disaggregation \\
& DAE~\cite{jia2019matrix} & Denoising autoencoder & Signal reconstruction & Disaggregation \\
& BERT+~\cite{li2024bert} & Bert Blocks & Ultilize extra user embeddings  & Disaggregation\\
& UNetNILM~\cite{faustine2020unet} & Skip-connected encoder-decoder & Multi-task learning & Both \\
\bottomrule
\multicolumn{5}{l}{\textit{\footnotesize{*Tasks include \textbf{State Recognition} (appliance state recognition), \textbf{Disaggregation} (energy injection disaggregation), and \textbf{Both} (both tasks).}}}
\end{tabular}}
\end{table*}

For fair comparison and evaluation, we select both traditional NILM methods, such as FHMM, and modern deep learning approaches. Although some methods specialize in appliance state recognition or energy injection disaggregation, methods such as FHMM and UNetNILM handle both tasks simultaneously. We additionally benchmark against recent BTM-specific approaches including SunDance~\cite{chen2017sundance} for solar disaggregation and BERT+~\cite{li2024bert} for injection estimation. This diverse selection enables comprehensive evaluation of \textsf{DualNILM}'s performance across different technical approaches and computational paradigms, with particular attention to the unique challenges posed by BTM energy injection.

Detailed configurations and implementation specifics are provided in Appendix~\ref{appendix:benchmarks}. All benchmark models were implemented with carefully tuned hyperparameters to ensure fair comparison. This comprehensive selection enables evaluation of different technical approaches, with particular attention to methods capable of handling BTM energy injection.

\section{Experimental Results}\label{sec:results}

\begin{figure}[t]
    \centering
    \includegraphics[width=1.00\linewidth, trim={0.2cm 0 0cm 0}, clip]{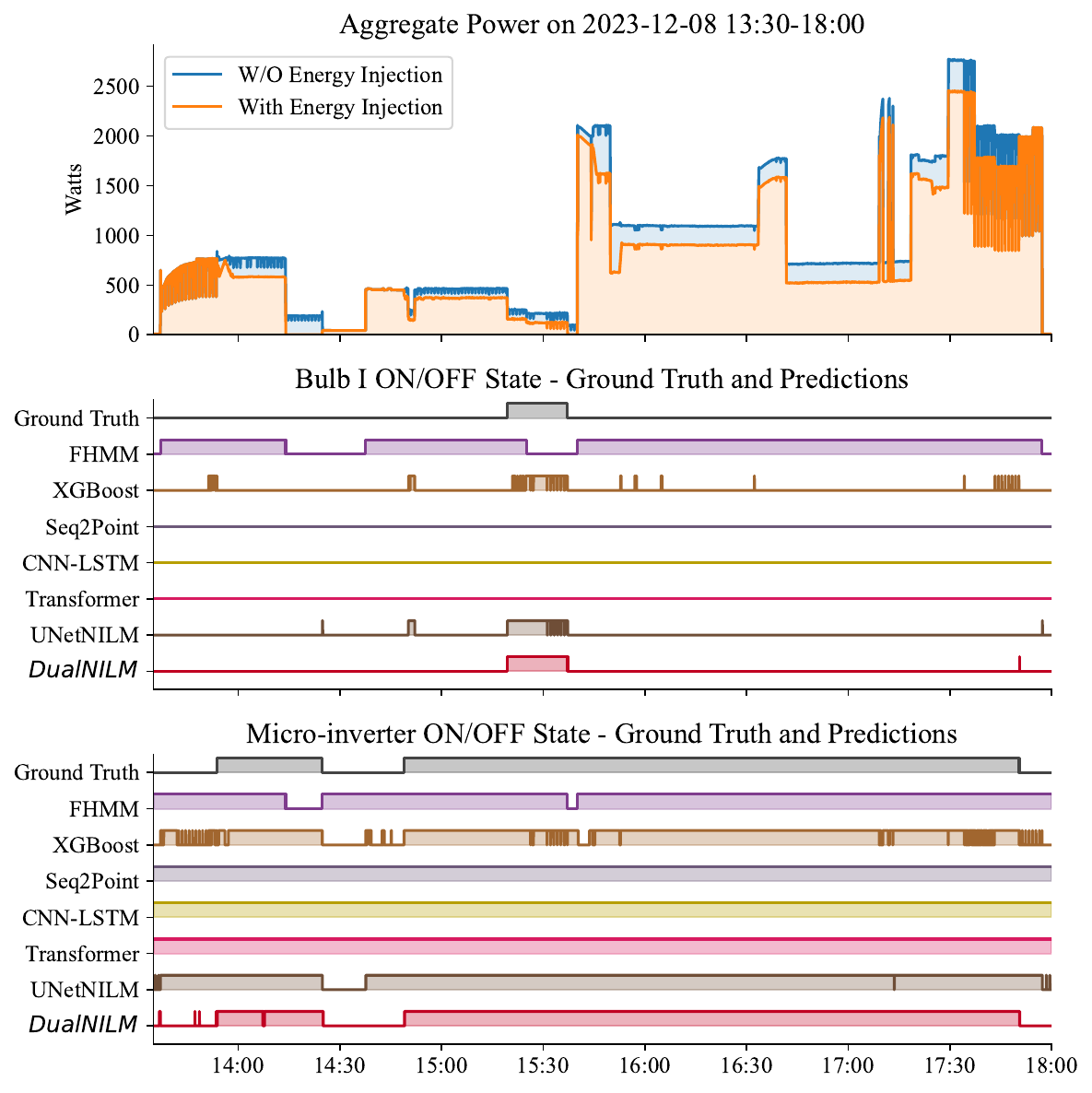}
    \caption{Visualization of Bulb I and Micro-inverter State's Ground Truth and Predictions across All Benchmark Methods on Dec.8, 2023 of our laboratory data.}
    \label{fig:1208model_comparison}
    \vspace{-0.3cm}
\end{figure}

This section presents a comprehensive evaluation of \textsf{DualNILM} across real-world laboratory measurements and synthetic public datasets augmented with PV injection. We analyze performance on controlled micro-inverter injection collected in our laboratory and validate generalization through synthetic augmentation of REDD and UK-DALE datasets, demonstrating the framework's effectiveness across diverse BTM injection scenarios.

\begin{table*}[h!]
\centering
\caption{Overall Avg. Performance Metrics of Appliance State Recognition (Cross-validation on all splits of each Dataset)}\label{tab:total_overall_avg}
\vspace{-2pt}
\resizebox{0.70\textwidth}{!}{
\setlength{\tabcolsep}{1.2mm}
\renewcommand{\arraystretch}{0.97}
\newcommand{\meanstd}[2]{#1$_{\scriptstyle\,\pm#2}$}
\begin{tabular}{c|c|c|cccc}
\hline
\multicolumn{2}{c|}{\makecell[c]{Dataset}} & \makecell[c]{Method} & Accuracy~$\uparrow$(\%) & Recall~$\uparrow$(\%) & Precision~$\uparrow$(\%) & F1~$\uparrow$(\%) \\
\hline
\multicolumn{2}{c|}{\multirow{7}{*}{\centering \makecell[c]{Laboratory Dataset \\ (Real Injection)}}} & FHMM & \meanstd{60.95}{27.96} & \meanstd{28.32}{40.81} & \meanstd{16.84}{28.20} & \meanstd{17.25}{27.23} \\
\multicolumn{2}{c|}{} & XGBoost & \meanstd{85.82}{2.07} & \meanstd{48.20}{7.69} & \meanstd{52.10}{6.02} & \meanstd{47.75}{6.75} \\
\multicolumn{2}{c|}{} & Seq2Point & \meanstd{67.63}{17.08} & \meanstd{53.06}{10.80} & \meanstd{27.86}{14.16} & \meanstd{34.27}{13.29} \\
\multicolumn{2}{c|}{} & CNN-LSTM & \meanstd{74.03}{15.51} & \meanstd{52.97}{11.02} & \meanstd{30.29}{14.28} & \meanstd{36.67}{13.25} \\
\multicolumn{2}{c|}{} & Transformer & \meanstd{81.79}{12.65} & \meanstd{52.91}{10.76} & \meanstd{38.16}{15.53} & \meanstd{42.76}{14.02} \\  
\multicolumn{2}{c|}{} & UNetNILM & \meanstd{89.20}{9.76} & \meanstd{54.01}{27.35} & \meanstd{60.81}{30.96} & \meanstd{53.89}{27.00} \\
\multicolumn{2}{c|}{} & \textbf{\ouralg} & \textbf{\meanstd{98.54}{1.71}} & \textbf{\meanstd{85.43}{28.03}} & \textbf{\meanstd{84.16}{28.45}} & \textbf{\meanstd{84.54}{27.93}} \\
\hline
\multirow{21}{*}{\shortstack{REDD\\with\\Synthetic\\Injection}}
& \multirow{7}{*}{House 1} & FHMM & \meanstd{79.52}{39.21} & \meanstd{1.46}{3.60} & \meanstd{13.47}{33.46} & \meanstd{2.07}{6.20} \\
& & XGBoost & \meanstd{97.90}{0.38} & \meanstd{74.50}{2.64} & \meanstd{88.74}{2.08} & \meanstd{79.18}{1.79} \\
& & Seq2Point & \meanstd{97.65}{0.46} & \meanstd{42.20}{3.43} & \meanstd{56.42}{1.16} & \meanstd{46.07}{4.03} \\
& & CNN-LSTM & \meanstd{98.67}{0.46} & \meanstd{46.32}{2.41} & \meanstd{50.75}{3.19} & \meanstd{47.96}{2.07} \\
& & Transformer & \meanstd{95.99}{1.10} & \meanstd{56.08}{5.37} & \meanstd{54.10}{8.04} & \meanstd{53.42}{6.31} \\
& & UNetNILM &  \meanstd{99.05}{0.31} & \meanstd{76.11}{3.42} & \meanstd{85.22}{5.99} & \meanstd{75.27}{4.25} \\
& & \textbf{\ouralg} & \textbf{\meanstd{99.84}{0.07}} & \textbf{\meanstd{96.26}{3.15}} & \textbf{\meanstd{97.11}{1.90}} & \textbf{\meanstd{96.64}{2.42}} \\
\cline{2-7}
& \multirow{7}{*}{House 2}& FHMM        & \meanstd{75.08}{42.51}        & \meanstd{0.58}{1.35}          & \meanstd{0.34}{0.79}          & \meanstd{0.43}{1.00}          \\
&  & XGBoost     & \meanstd{98.43}{0.09}         & \meanstd{92.47}{3.94}         & \textbf{\meanstd{95.58}{1.21}} & \textbf{\meanstd{93.53}{3.15}} \\
&  & Seq2Point   & \meanstd{91.43}{1.50}         & \meanstd{52.90}{2.47}         & \meanstd{58.44}{1.59}         & \meanstd{50.85}{2.27}         \\
&  & CNN-LSTM    & \meanstd{98.86}{0.32}         & \meanstd{49.19}{0.39}         & \meanstd{48.74}{0.15}         & \meanstd{48.96}{0.26}         \\
&  & Transformer & \meanstd{97.92}{0.19}         & \meanstd{47.78}{0.35}         & \meanstd{47.98}{0.50}         & \meanstd{47.87}{0.22}         \\
&  & UNetNILM    & \meanstd{99.57}{0.22}         & \meanstd{64.16}{16.03}        & \meanstd{77.22}{20.61}        & \meanstd{67.28}{17.02}        \\
&  & \textbf{\ouralg}   & \textbf{\meanstd{99.70}{0.22}} & \textbf{\meanstd{92.48}{5.55}} & \meanstd{94.00}{5.47}         & \meanstd{93.11}{4.96}         \\
\cline{2-7}
& \multirow{7}{*}{House 3} & FHMM        & \meanstd{83.84}{27.80}          & \meanstd{4.53}{8.79}          & \meanstd{0.90}{1.69}           & \meanstd{1.49}{2.80}           \\
& & XGBoost     & \meanstd{94.89}{2.15}          & \meanstd{46.22}{0.48}         & \meanstd{56.38}{2.49}          & \meanstd{50.51}{1.28}          \\
& & Seq2Point   & \meanstd{98.10}{0.88}          & \meanstd{42.78}{7.36}         & \meanstd{50.90}{11.98}         & \meanstd{44.47}{7.97}          \\
& & CNN-LSTM    & \meanstd{95.92}{2.70}          & \meanstd{28.85}{5.34}         & \meanstd{38.17}{0.75}          & \meanstd{32.62}{3.48}          \\
& & Transformer & \meanstd{97.81}{0.61}          & \meanstd{40.21}{5.32}         & \meanstd{38.89}{4.60}          & \meanstd{39.43}{4.67}          \\
& & UNetNILM    & \meanstd{99.07}{0.32}          & \meanstd{59.91}{0.04}         & \meanstd{54.73}{1.56}          & \meanstd{57.10}{0.95}          \\
& & \textbf{\ouralg}    & \textbf{\meanstd{99.69}{0.24}} & \textbf{\meanstd{77.42}{17.13}}& \textbf{\meanstd{84.46}{12.18}} & \textbf{\meanstd{80.04}{15.51}} \\
\hline
\multirow{14}{*}{\shortstack{UKDALE\\with\\Synthetic\\Injection}}& \multirow{7}{*}{House 1} & FHMM         & \meanstd{89.25}{23.05}          & \meanstd{1.58}{3.57}           & \meanstd{0.76}{1.87}            & \meanstd{1.01}{2.39}            \\
& & XGBoost      & \meanstd{93.64}{0.42}           & \meanstd{41.74}{1.40}          & \meanstd{68.04}{3.98}           & \meanstd{45.97}{2.56}           \\
& & Seq2Point    & \meanstd{88.70}{0.58}           & \meanstd{41.47}{2.83}          & \meanstd{43.29}{5.22}           & \meanstd{36.79}{3.32}           \\
& & CNN-LSTM     & \meanstd{98.33}{0.24}           & \meanstd{31.76}{0.20}          & \meanstd{31.29}{0.24}           & \meanstd{31.52}{0.15}           \\
& & Transformer  & \meanstd{96.90}{0.29}           & \meanstd{33.65}{1.03}          & \meanstd{27.76}{2.49}           & \meanstd{30.26}{1.69}           \\
& & UNetNILM     & \meanstd{98.76}{0.16}           & \meanstd{52.03}{5.14}          & \meanstd{72.01}{18.92}          & \meanstd{55.20}{7.02}           \\
& & \textbf{\ouralg}
                & \textbf{\meanstd{99.72}{0.21}}  & \textbf{\meanstd{95.23}{3.53}} & \textbf{\meanstd{94.30}{4.09}}  & \textbf{\meanstd{94.71}{3.63}}  \\
\cline{2-7}
& \multirow{7}{*}{House 2} & FHMM         & \meanstd{82.95}{36.49}          & \meanstd{0.62}{1.17}           & \meanstd{0.52}{1.01}            & \meanstd{0.55}{1.06}            \\
& & XGBoost      & \meanstd{96.13}{0.61}          & \meanstd{66.61}{5.32}          & \meanstd{79.19}{9.68}           & \meanstd{70.67}{6.81}           \\
& & Seq2Point    & \meanstd{98.83}{0.18}          & \meanstd{42.68}{3.09}          & \meanstd{46.49}{12.53}          & \meanstd{41.92}{3.19}           \\
& & CNN-LSTM     & \meanstd{97.81}{0.86}          & \meanstd{41.65}{5.05}          & \meanstd{45.48}{3.20}           & \meanstd{43.33}{4.21}           \\
& & Transformer  & \meanstd{99.13}{0.14}          & \meanstd{59.88}{5.10}          & \meanstd{77.44}{7.86}           & \meanstd{61.73}{4.66}           \\
& & UNetNILM     & \meanstd{99.17}{0.19}          & \meanstd{64.25}{5.53}          & \meanstd{68.74}{5.74}           & \meanstd{65.43}{4.72}           \\
& & \textbf{\ouralg}
                & \textbf{\meanstd{99.73}{0.04}}  & \textbf{\meanstd{90.63}{5.01}} & \textbf{\meanstd{94.46}{3.36}}  & \textbf{\meanstd{92.31}{3.76}}  \\
\bottomrule
\end{tabular}}
\vspace{-5pt}
\end{table*}

\begin{figure}[t]
    \centering
    \includegraphics[width=1.00\linewidth, trim={0.2cm 0 0cm 0}, clip]{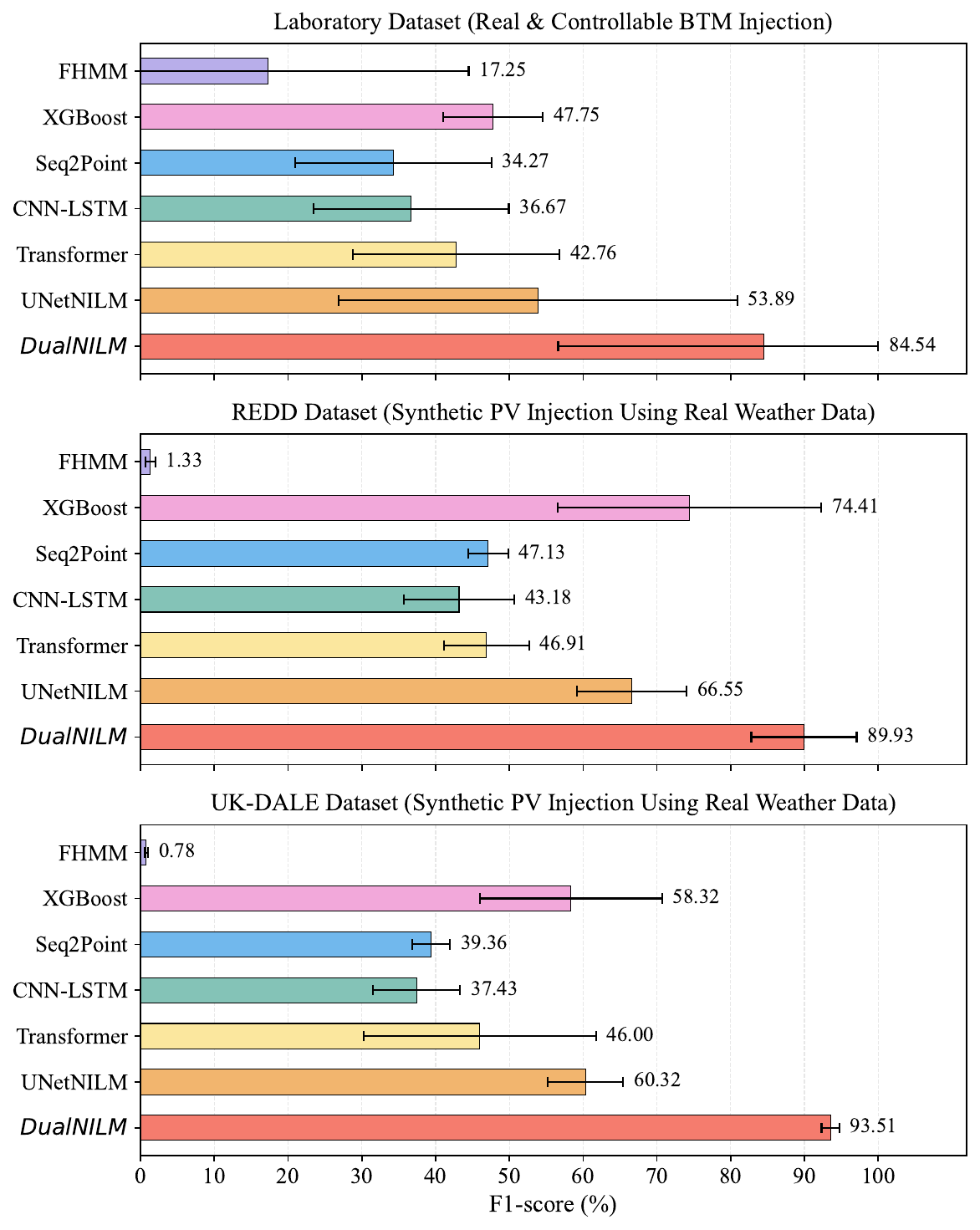}
    \caption{Comparison of overall average F1-scores for appliance state recognition across Laboratory, REDD, and UKDALE datasets.}
    \label{fig:f1_comparison}
\end{figure}

\paragraph{\textbf{Overview.}} Tables~\ref{tab:total_overall_avg} and \ref{tab:total_overall_digg_avg} report comprehensive performance metrics for appliance-state recognition and energy-injection disaggregation. We evaluated them in two complementary settings: i) controlled laboratory experiments using a real 500~W micro-inverter as the BTM injection; and ii) household datasets (REDD and UK-DALE) augmented with synthetic PV profiles to emulate typical residential consumption–generation patterns. This multi-faceted design enables robust validation across diverse, realistic household usage regimes. Figure~\ref{fig:f1_comparison} shows the Avg. F1-score performance comparison across all datasets.

\subsection{Performance on Laboratory Dataset}
\label{subsec:lab-performance}

\paragraph{\textbf{Appliance State Recognition.}}
The laboratory experiments provide critical insights into real-world performance with actual hardware effects. Table~\ref{tab:total_overall_avg} (first block) reveals fundamental challenges that bidirectional power flows pose to conventional NILM methods. FHMM, despite its theoretical ability to model multiple hidden states, achieves only 17.25\% F1-score. This poor performance stems from its factorial hidden Markov model structure, which assumes discrete state transitions and cannot accommodate the continuous power variations introduced by inverter operation. The model essentially interprets injection-induced fluctuations as rapid appliance state changes, leading to excessive false positives.

Among deep learning approaches, we observe a clear performance gradient correlating with architectural complexity. Seq2Point, designed for point-wise load disaggregation, achieves an average score of 34.27\% F1 as its narrow temporal window cannot distinguish between appliance transitions and injection variations. CNN-LSTM improves to 36.67\% by incorporating temporal memory, while the Transformer reaches 42.76\% through self-attention mechanisms that capture longer-range dependencies. UNetNILM, with its U-Net architecture originally designed for multi-task NILM, achieves 53.89\% F1-score by jointly modeling states and power consumption, though it lacks explicit injection modeling.

\textsf{DualNILM} achieves 84.54\% F1-score, representing a 57\% relative improvement over the best baseline. This dramatic improvement validates our dual-task hypothesis: explicitly modeling injection as a parallel objective enables the network to disambiguate genuine appliance activities from generation-induced variations. Notably, XGBoost achieves 47.75\% F1-score through its tree-based ensemble learning, which implicitly discovers patterns in the dual-channel input. However, unlike \ouralg, it cannot provide explicit injection estimates, and its performance varies significantly across different scenarios. Figure~\ref{fig:1208model_comparison} visualizes this capability, showing accurate state detection even during concurrent inverter operation periods.

\paragraph{\textbf{Energy Injection Disaggregation.}}
For continuous injection estimation (Figure~\ref{fig:rmse_comparison} and  Appendix Table~\ref{tab:total_overall_digg_avg}), the laboratory results reveal interesting trade-offs between different regression approaches. Seq2Seq achieves competitive MAE (0.2622) through its sequence-to-sequence architecture optimized for time series prediction. However, its higher RMSE (0.3512) indicates susceptibility to outliers during rapid injection transitions. DAE shows the poorest performance (RMSE: 0.8830), as its denoising objective is misaligned with the injection estimation task. UNetNILM demonstrates reasonable performance (RMSE: 0.2305) by leveraging its multiscale feature extraction.

\textsf{DualNILM} achieves superior overall performance, yielding average RMSE and MAE of 0.1429 and 0.0804, respectively. The superior accuracy stems from the synergistic relationship between tasks: accurate appliance state detection helps refine injection estimates by accounting for consumption changes, while precise injection estimation prevents false appliance detections during generation fluctuations.

\subsection{Performance on Synthetic Public Datasets}
\label{subsec:synthetic-performance}

The synthetic experiments validate generalization across diverse household types and consumption patterns. We augment REDD and UK-DALE with realistic PV generation profiles derived from solar irradiance data, creating challenging scenarios where generation can exceed baseload consumption during peak hours. Figure~\ref {fig:ukdale1_simulation_demo} reveals a challenging disaggregation scenario on UKDALE House 1 during midday hours, where only \ouralg successfully captures both the refrigerator's cycling pattern and the microwave's short operation compared with other benchmark methods.

\paragraph{\textbf{Performance of Traditional NILM Approaches.}}
\begin{itemize}[leftmargin=*, topsep=0pt, itemsep=2pt]
\item \textbf{FHMM} demonstrates severe degradation on both datasets, with F1-scores below 3\% on most houses. This catastrophic failure occurs because the factorial model cannot handle negative net power when generation exceeds consumption. The hidden states, designed to represent appliance ON/OFF combinations, become meaningless when the observation can be negative.

\item \textbf{XGBoost}, as a gradient boosting ensemble method, shows interesting dataset-dependent behavior. On REDD, it maintains reasonable performance (F1-scores: 79.18\% House 1, 93.53\% House 2) through handcrafted features including time-of-day, power derivatives, and statistical aggregates. However, it degrades significantly on UK-DALE (F1-scores: 45.97\% House 1, 70.67\% House 2), indicating sensitivity to dataset-specific appliance distributions and consumption patterns. This highlights a fundamental limitation of traditional machine learning: the need for manual feature design that may not generalize across different geographical contexts.
\paragraph{\textbf{Deep Learning Methods Performance Analysis.}}
Deep learning methods exhibit varied resilience to PV injection based on their architectural inductive biases:
\end{itemize}
\begin{itemize}[leftmargin=*, topsep=0pt, itemsep=2pt]
    \item \textbf{Seq2Point} maintains high accuracy ($>88\%$) but suffers from precision-recall imbalance. Its point-wise prediction cannot leverage temporal context to distinguish injection from load reduction, resulting in conservative predictions biased toward OFF states (F1-scores: 36-50\%).
    
    \item \textbf{CNN-LSTM} shows similar patterns with slightly better recall due to LSTM memory, but still achieves F1-scores below 50\% on most houses. The convolutional features capture local patterns but lack the global context needed for injection disambiguation.
    
    \item \textbf{Transformer} demonstrates improved resilience (F1-scores: 30-61\%) through self-attention, which can correlate time-of-day with typical solar generation patterns. However, without explicit injection modeling, it still confuses cloud-induced generation drops with appliance activations.
    
    \item \textbf{UNetNILM} exhibits high variance across houses, F1-scores from 57.10\% (REDD House 3) to 75.27\% (REDD House 1). This instability suggests difficulty balancing its dual objectives of state classification and power regression when injection is present but not explicitly modeled.
\end{itemize}

\begin{figure}[t]
    \centering
    \includegraphics[width=1.00\linewidth, trim={0.2cm 0 0cm 0}, clip]{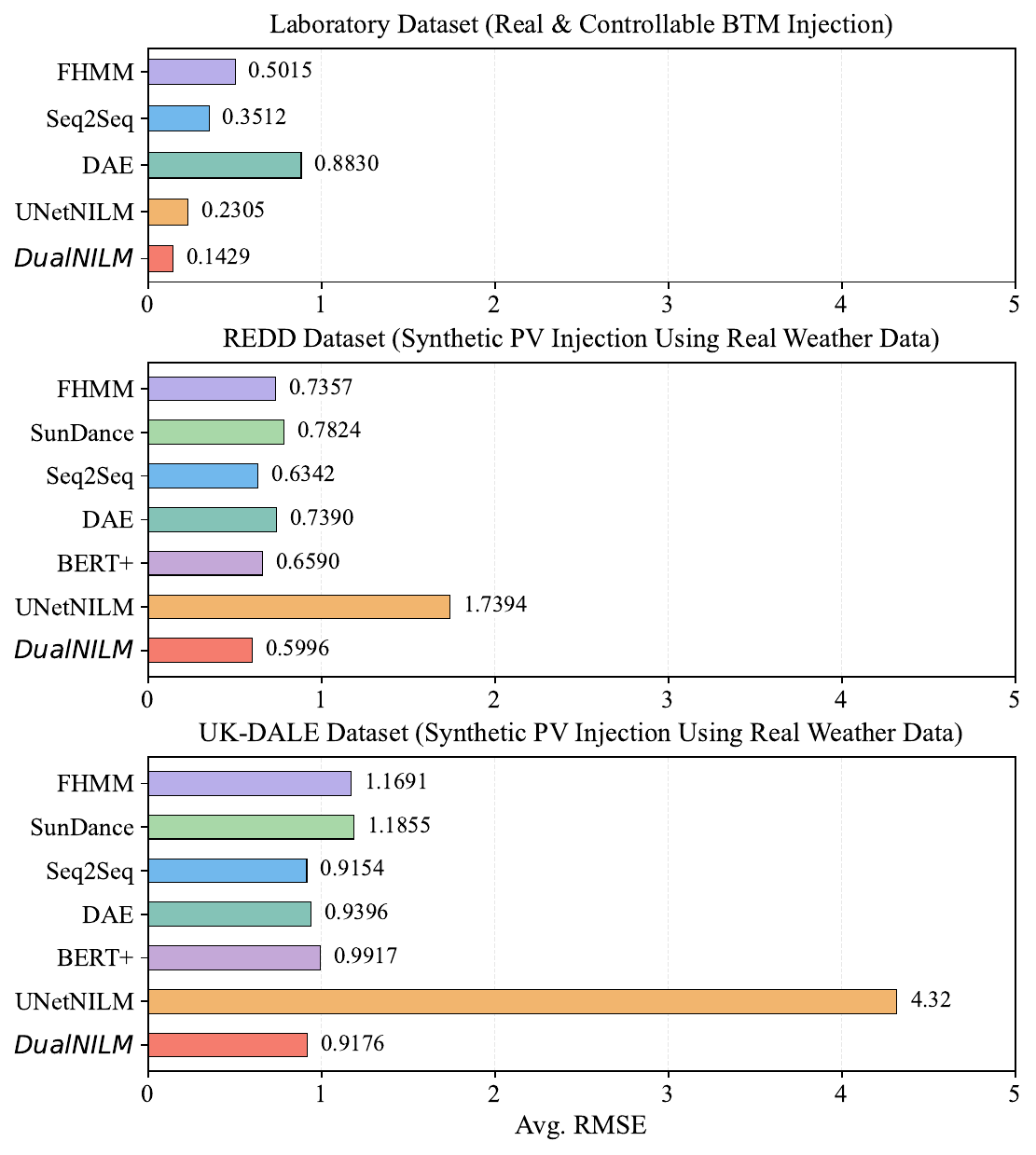}
    \caption{Comparison of overall average RMSE for BTM energy injection disaggregation across Laboratory, REDD, and UKDALE datasets. See full results in Table~\ref{tab:total_overall_digg_avg}.}
    \label{fig:rmse_comparison}
\end{figure}

\paragraph{\textbf{DualNILM Performance Consistency.}}
\textsf{DualNILM} maintains robust performance across all test scenarios. On REDD, it achieves F1-scores of 96.64\%, 93.11\%, and 80.04\% for Houses 1-3, respectively. On UK-DALE, it achieves 94.71\% and 92.31\% for Houses 1-2. This consistency across geographically diverse datasets with different appliance types, usage patterns, and synthetic injection profiles demonstrates genuine generalization capability.

The slightly lower performance on REDD House 3 (F1-score 80.04\%) provides valuable insights. This house contains more low-power appliances whose signatures are easily masked by injection variations. Yet \textsf{DualNILM} still substantially outperforms all baselines, validating the robustness of the dual-task approach even in challenging scenarios. The performance variance of XGBoost across REDD houses (50.51\% to 93.53\% F1-score) and its degraded performance on UK-DALE (45.97\% on House 1) suggest that tree-based pattern learning, while effective in specific scenarios, lacks the generalization capability of explicit multi-task modeling.

\paragraph{\textbf{Injection Disaggregation Across Datasets.}}
Figure~\ref{fig:rmse_comparison} and Table~\ref{tab:total_overall_digg_avg} in Appendix reveal that injection estimation becomes more challenging on public datasets compared to laboratory settings. This is expected as the synthetic injection exhibits more complex temporal dynamics including cloud transients and seasonal variations. Seq2Seq occasionally achieves competitive RMSE on individual houses but shows less consistent MAE performance. SunDance~\cite{chen2017sundance}, designed specifically for solar disaggregation using physical generation models, exhibits limited performance compared with other NILM disaggregation baselines, while BERT+ demonstrates competitive performance (REDD: 0.6590 average RMSE; UK-DALE: 0.9919 average RMSE). Notably, UNetNILM exhibits severe instability on UK-DALE House 1 (RMSE: 7.5014 in full Table~\ref{tab:total_overall_digg_avg}), suggesting training convergence issues when the injection magnitude varies widely. This instability, contrasted with its reasonable performance on REDD datasets, highlights the sensitivity of standard multi-task architectures to dataset-specific characteristics without proper architectural design for handling bidirectional power flows.
\textsf{DualNILM} delivers balanced performance across all metrics and houses. While not always achieving the absolute lowest RMSE on every house, it consistently ranks among the top performers for both RMSE and MAE, demonstrating robust generalization. The multi-task learning enables better handling of the complex interaction between varying consumption patterns and injection profiles.

\subsection{Ablation Study on \ouralg}\label{sec:ablation}

To validate our architectural design choices, we conducted comprehensive ablation experiments examining joint versus separate training strategies and input feature configurations. Results in Table~\ref{tab:ablation_joint} demonstrate that joint optimization improves appliance state recognition F1-score by 6.44\% over separate state-only training (88.27\% $\rightarrow$ 94.71\%) and reduces injection estimation RMSE by 5.4\% compared to separate injection-only training (0.9521 $\rightarrow$ 0.9008). The synergistic relationship between tasks enables the model to leverage accurate state predictions to refine injection estimates while using injection awareness to prevent false appliance detections during generation fluctuations.

\begin{figure}[t]
    \centering
    \includegraphics[width=0.95\linewidth]{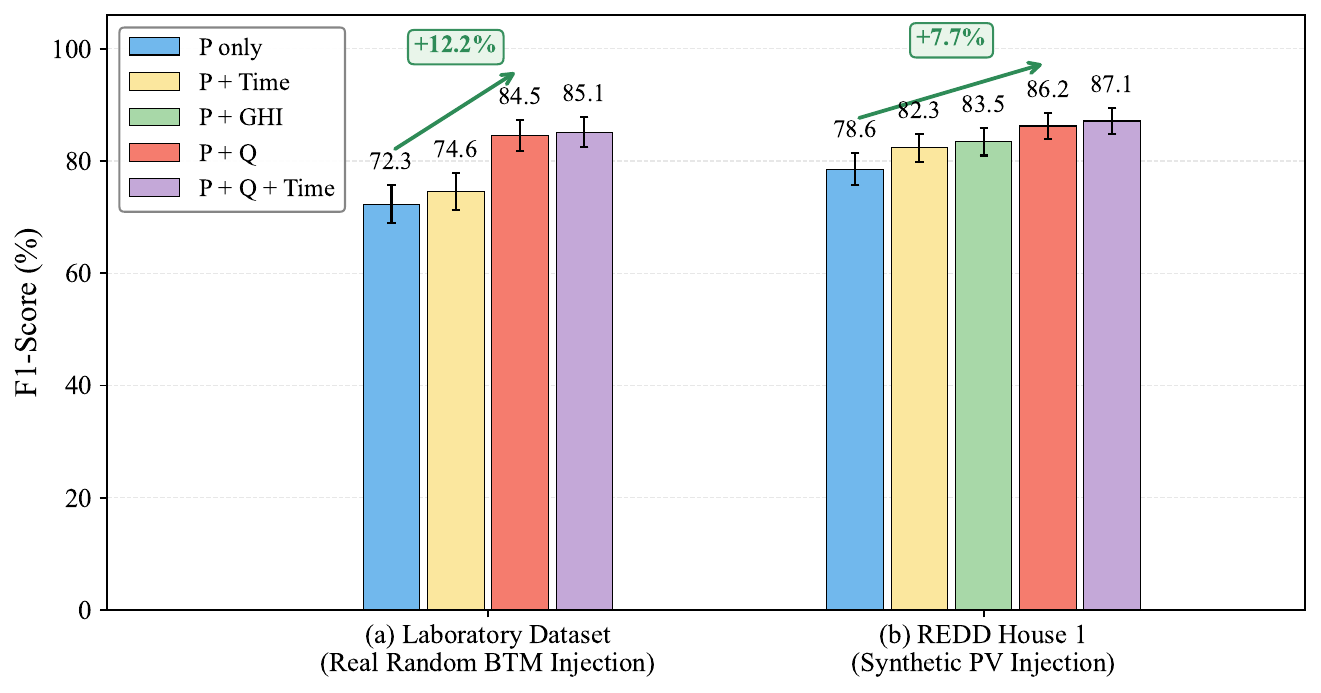}
    \caption{Feature ablation F1-score comparison on Laboratory and REDD House 1 datasets. With Reactive power (P+Q) provides the largest performance gain, validating that appliance-specific reactive signatures remain discriminative under BTM injection.}
    \label{fig:feature_ablation_f1}
    \vspace{-0.5cm}
\end{figure}

Feature ablation studies (Figure~\ref{fig:feature_ablation_f1}) reveal that reactive power provides critical discriminative information. On laboratory data with random injection patterns, adding reactive power to active power improves F1-score by 12.23\% (72.31\% $\rightarrow$ 84.54\%), validating our insight that inverter near-unity power factor operation preserves appliance-specific reactive signatures. While external features such as weather data and time encoding provide marginal improvements on synthetic PV datasets, the core active and reactive power~(P+Q) configuration achieves robust performance without communication overhead, enabling practical edge deployment. Complete ablation results and detailed analysis are provided in Appendix~\ref{app:ablation}.

\subsection{Key Insights and Analysis}
\label{subsec:insights}

Several critical insights emerge from the comprehensive evaluation:

\paragraph{\textbf{Importance of Explicit Injection Modeling.}}
The performance gap between \textsf{DualNILM} and UNetNILM, both multi-task architectures, underscores the importance of explicitly modeling injection rather than treating it as a general regression target. Although UNetNILM jointly performs classification and regression, it lacks the architectural separation and cross-task synergy that enables \textsf{DualNILM} to learn different, implicitly connected representations for consumption and generation.

\paragraph{\textbf{Complementary Value of Real and Synthetic Data.}}
Laboratory experiments capture real hardware effects, including inverter switching noise, electromagnetic interference, and actual appliance transients that are difficult to simulate. In contrast, synthetic datasets provide diversity in household types, appliance portfolios, and usage patterns essential for evaluating generalization. Together, they provide comprehensive validation that neither could alone achieve.

\paragraph{\textbf{Method-Specific Failure Modes and Optimization Challenges.}}
Our results reveal distinct failure patterns that illuminate the fundamental challenges of BTM injection. 

Traditional methods, such as FHMM, fail due to violated mathematical assumptions, and their factorial structure assumes discrete states incompatible with continuous injection variations. Notably, XGBoost serves as a strong baseline that achieves competitive performance (74.41\% average F1 on REDD from Table~\ref{tab:redd_overall_avg_2kw}) through implicit feature engineering via recursive tree splits on the dual-channel input. However, it is fundamentally limited by its inability to provide explicit injection estimates and its reliance on pattern memorization, which makes it susceptible to distributional shifts.

Deep learning methods encounter more complex optimization challenges. The introduction of the reactive power channel and multiple appliance types with imbalanced activation patterns significantly expands the optimization landscape. Seq2Point (34.27\% F1 on laboratory data) cannot disambiguate temporal patterns due to its narrow window. CNN-LSTM and Transformer architectures, despite sophisticated temporal modeling, struggle with convergence since the simultaneous presence of sparse appliance activations and continuous injection creates conflicting gradient signals. These diverse failure modes validate that the injection challenge requires fundamental architectural innovation, not incremental improvements.

\paragraph{\textbf{Practical Deployment Implications.}}
The results have immediate practical implications for the deployment of smart grids. Existing NILM systems will experience severe degradation in solar or BTM generation-equipped homes, particularly during peak generation hours. Our analysis shows that performance degradation is not uniform but depends critically on the relationship between injection magnitude and base load levels. When injection exceeds 40\% of typical consumption, even sophisticated baselines fail to maintain acceptable accuracy. The consistent performance of \textsf{DualNILM} across diverse scenarios suggests it could be deployed as a software upgrade to existing smart meter infrastructure, enabling continued load monitoring despite distributed generation.

\section{Conclusion}\label{sec:conclusion}
We presented \textsf{DualNILM}, a deep multi-task learning framework tailored to address the challenges of NILM in the presence of BTM energy injection, such as micro-inverters and photovoltaic system outputs. Extensive experiments on real-world laboratory datasets with micro-inverter injection, as well as synthesized PV injection datasets derived from well-known NILM benchmarks, demonstrate \textsf{DualNILM}'s superior performance and the strong potential for advancing NILM solutions in modern energy systems with BTM energy injection.

Our findings underscore~\textsf{DualNILM}'s potential for robust NILM in modern energy systems that integrate renewable energy sources, and our provided open research infrastructure~\href{https://github.com/MathAdventurer/PV-Augmented-NILM-Datasets}{[\textit{link}]} will help the NILM community continue to advance research in this critical direction.

\bibliographystyle{ACM-Reference-Format}
\bibliography{reference}

\newpage
\onecolumn
\appendix
\section*{Appendix}

\section*{Ethical Considerations}
\label{sec:ethics}

This research adheres to established ethical principles for computational research and poses minimal ethical risks. Our study involves purely technical methodology development for energy disaggregation systems using laboratory-controlled measurements and synthetic datasets derived from publicly available benchmarks (REDD, UK-DALE).

\textbf{Data Sources and Privacy.} All experimental data consists of: (1) controlled laboratory measurements from our own micro-inverter setup, and (2) synthetic photovoltaic injection patterns generated using established solar modeling techniques without personal information. No human subjects, user behavioral data, or personally identifiable information were involved in this research.

\textbf{Societal Impact.} The proposed \textsf{DualNILM} framework aims to enhance energy monitoring capabilities in renewable-integrated smart grids, contributing to more efficient energy management and supporting the transition to sustainable energy systems. The method addresses a purely technical challenge without political, religious, or culturally sensitive implications.

\textbf{Open Science.} To promote reproducibility and advance community research, we provide open access to our synthetic PV-augmented datasets and benchmark implementations, following best practices for responsible data sharing in the energy informatics community.

\appendix
\section{Theoretical Analysis of the Challenge: When Negative Flows Disrupt Classical NILM}
\label{appendix:challenge_analysis}

In the classic NILM framework, the observed signal $\mathbf{y} \in \mathbb{R}^T$ (where $T$ is the number of time samples) is seen as a single-channel mixture of multiple component signals (appliances), each assumed to be nonnegative over time. When BTM injection is introduced, at least one component can take negative values, increasing the dimensionality of the solution space and weakening certain identifiability properties that hold under nonnegativity constraints.

\subsection{Matrix Formulation of NILM}

Consider $N$ appliances in a building (including the "injection appliance"). For each appliance $n \in \{1,\dots,N\}$, define:
\[
\mathbf{x}_{n} = (x_{n}(1), x_{n}(2),\ldots, x_{n}(T))^\top \in \mathbb{R}^{T},
\]
where $x_{n}(t)$ is the instantaneous power consumption (or production, if negative). A binary activation sequence $\mathbf{s}_{n} \in \{0,1\}^{T}$ indicates whether appliance $n$ is operating at each time $t$. Thus, the aggregate measured load $\mathbf{y}\in\mathbb{R}^{T}$ satisfies:
\begin{equation}
\label{eq:matrix-form}
\mathbf{y} = \sum_{n=1}^{N} (\mathbf{s}_{n} \odot \mathbf{x}_{n}) + \boldsymbol{\epsilon},
\end{equation}
where $\odot$ denotes the elementwise (Hadamard) product and $\boldsymbol{\epsilon}\in\mathbb{R}^T$ is measurement noise. In classical NILM, it is assumed that $\mathbf{x}_{n}(t)\geq 0$ for every $n$ and $t$, reflecting purely consumptive loads. This naturally constrains the solution space to a nonnegative cone, aiding disaggregation methods that leverage structure from Nonnegative Matrix Factorization (NMF) or other constrained dictionary approaches.

\subsection{Why BTM Injection Complicates NILM}

In the language of signal processing and blind source separation (BSS), classical NILM can be viewed as a single-channel mixture of purely nonnegative sources. The geometry of such mixtures is conic: if each device's load lies in the nonnegative orthant, standard factorization approaches (e.g., nonnegative matrix factorization) can leverage convexity to help isolate distinct appliance patterns~\cite{lee1999learningNMF}.

Once one or more sources are allowed to take negative values (e.g., exporting solar power), the total load no longer resides in the same conic space. Mathematically, the feasible solution set expands:
\[
\mathcal{F}_{\text{injection}} = \left\{
(\mathbf{s}, \mathbf{x}):
\mathbf{x}_n\in\mathbb{R}_+^T \text{ for } n \neq k,\, 
\mathbf{x}_k \in \mathbb{R}^T,\, 
\mathbf{y} = \sum_{n=1}^{N-1} (\mathbf{s}_n\odot \mathbf{x}_n) + (\mathbf{s}_k\odot \mathbf{x}_k)
\right\}.
\]

Because one column (the injection device) spans a broader region than the nonnegative orthant, various identifiability arguments from conic geometry and nonnegative matrix factorization lose traction. In essence:

\begin{itemize}
    \item \textbf{Larger Search Space:} With fewer constraints, multiple plausible ways to explain any observed dip or negative interval emerge, inflating ambiguity.
    \item \textbf{Weakened Identifiability:} Nonnegative factorizations often depend on convexity-based geometric constraints; negative entries undermine these constraints and can make standard disaggregation fall back on naive assumptions.
    \item \textbf{Low-Dimensional Data Shortcomings:} If the meter only provides a single power reading or a low-resolution signal, there is insufficient structure to separate negative flows from positive loads, compounding the difficulty.
\end{itemize}

\subsection{Relaxation to Negative Flows and Identifiability}

When behind-the-meter solar (or other storage/generation) is present, the corresponding $\mathbf{x}_{k}$ for that source may take negative values at certain time points:
\[
x_{k}(t) < 0 \quad \text{(exporting energy to the grid at time $t$)}.
\]
Hence, at least one column in the set $\{\mathbf{x}_{1},\ldots,\mathbf{x}_{N}\}$ is no longer restricted to $\mathbb{R}_{\geq 0}^T$ but spans $\mathbb{R}^T$, effectively dilating the feasible solution set as shown in Equation~\ref{eq:matrix-form}.

\textbf{Degraded Identifiability and Solution Stability.} Under classical NILM assumptions (fully nonnegative), more stable decompositions can sometimes be achieved when the underlying sources have distinct support or exhibit certain "sparse plus low-rank" structures. These identifiability arguments are often reminiscent of theorems in blind source separation (BSS) and NMF, where the nonnegative cone enforces a geometric structure that "pins down" solutions. Once negative values are allowed (even for a single source), the problem becomes closer to an unconstrained BSS, for which solution identifiability is well known to deteriorate significantly. In more formal terms, NMF-based identifiability relies on the fact that data points lie in a convex cone generated by the basis columns~\cite{lee1999learningNMF}; introducing negative components breaks those convex cone arguments.

\subsection{Signal Space Interpretation and Rank Constraints}

To add further clarity, consider that we often (implicitly) factorize $\mathbf{y}$ over time into a linear combination of dictionary columns:
\begin{equation}
\label{eq:dictionary}
\mathbf{y} \approx \sum_{n=1}^N \mathbf{D}_n \boldsymbol{\alpha}_n, 
\end{equation}
where $\mathbf{D}_n \in \mathbb{R}^{T\times d_n}$ represents a small dictionary of "candidate appliance patterns" for device $n$, and $\boldsymbol{\alpha}_n \in \mathbb{R}^{d_n}$ are nonnegative coefficients that specify which pattern from $\mathbf{D}_n$ is used at each time. 

In a purely consumptive (classical) case, $\boldsymbol{\alpha}_n \succeq \mathbf{0}$ imposes a conic constraint. For BTM injection, the dictionary for the injecting device must allow negative entries. Thus, while the classical (nonnegative) scenario yields:
\[
\mathbf{y} \in \text{cone}(\mathbf{D}_1 \cup \dots \cup \mathbf{D}_N),
\]
the new scenario broadens the feasible set to:
\[
\mathbf{y} \in \text{span}(\mathbf{D}_{k}) + \text{cone}(\mathbf{D}_1\cup\dots\cup\mathbf{D}_{k-1}\cup \mathbf{D}_{k+1}\cup \dots \cup \mathbf{D}_N),
\]
where $\mathbf{D}_{k}$ is the dictionary for the injecting device $k$. Removing the nonnegativity restriction on that column effectively transforms the conic geometry into a partial subspace, increasing the rank of the set spanned by all feasible combinations. Consequently, some of the identifiability claims from nonnegative dictionary learning no longer apply.

In summary, allowing negative flows expands the solution space beyond the convex cone imposed by nonnegative appliances. This expansion degrades the ability to achieve stable solution identification and significantly increases the computational complexity of load disaggregation. Traditional NILM methods that treat all loads as nonnegative typically fail to disentangle consumption from behind-the-meter generation when measurements dip below zero or remain subdued for extended intervals, as demonstrated in our experimental results (Figure~\ref{fig:1208aggregate_power} and Table~\ref{tab:total_overall_avg}).

\section{Details of Model Design and Implementation}\label{appendix:modeldesign}

\subsection{Model Hyperparameters}

The overall architecture of \ouralg is illustrated in Figure~\ref{fig:DualNILM-architecture}. Our model comprises four main components: 1) CNN Encoders, 2) Transformer Encoders, 3) a Transformer Decoder, and 4) Task-specific projection layers. This design allows for the effective capture of both local and global temporal dependencies in the power consumption patterns while enabling multi-task learning.

Our \ouralg model is implemented with the following configurations:

\begin{itemize}
    \item \textbf{Input}: Sequences of length $T=300$ with $F=2$ features (active power and reactive power).
    \item \textbf{CNN Encoders}:
    \begin{itemize}
        \item Three 1D convolutional layers per feature, each with $C=64$ filters, kernel size of 5, and padding of 2.
        \item Each convolutional layer is followed by ReLU activation and Layer Normalization.
    \end{itemize}
    \item \textbf{Transformer Encoder}:
    \begin{itemize}
        \item One Transformer Encoder layer.
        \item Model dimension $D = F \times C = 128$.
        \item Number of heads: 8.
        \item Feedforward network dimension: 128.
        \item Layer Normalization applied after the encoder.
    \end{itemize}
    \item \textbf{Transformer Decoder}:
    \begin{itemize}
        \item One Transformer Decoder layer.
        \item Same model dimension and number of heads as the encoder.
    \end{itemize}
    \item \textbf{Task-Specific Projections}:
    \begin{itemize}
        \item For energy injection estimation: a linear layer mapping from $D$ to 256 units, followed by a ReLU activation, a dropout layer (dropout rate 0.2), another linear layer mapping to 1 unit, and a sigmoid activation.
        \item For appliance state detection: for each appliance, a linear layer mapping from $D$ to 1 unit, followed by a sigmoid activation.
    \end{itemize}
\end{itemize}

\section{Datasets and PV Simulation Details}\label{appendix:dataset}

\subsection{Simulated Datasets}
\subsubsection{Data Alignment and Resampling}

To integrate the solar data with the energy consumption data, we perform the following steps:

\begin{enumerate}
    \item \textbf{Time Zone Conversion:} All timestamps are converted to UTC to ensure temporal alignment.
    \item \textbf{Resampling:} The solar irradiance data is resampled to match the sampling rate of the consumption data (e.g., every 6 seconds for REDD).
    \item \textbf{Interpolation:} Missing values are filled using forward-fill and back-fill methods.
\end{enumerate}

\subsubsection{The Choice of Simulated 2kW PV Scale}\label{app:tinypvsurvey}

For the context of REDD and UKDALE, in the early-2010s PV market, residential new-construction PV systems most commonly fell in the 2-4~kW range~\cite{Barbose2014TrackingSunVII}, which puts our choice of  2~kW injection at the small-system end but still representative for that period.

From a siting perspective, NREL’s 2010 residential benchmark assumes $\sim$35~m$^{2}$ of array area using 27 modules (14.5\% efficiency) for a $\sim$5~kWdc system; by simple proportionality, a 2~kW array needs $\sim$14~m$^{2}$, well within the envelope of typical roof planes~\cite{Goodrich2012PVSystemPrices}.

Consistently, NREL’s rooftop technical-potential study requires just one contiguous roof plane of $\geq$10~m$^{2}$, noting this supports $\approx$1.5~kW at 15\% efficiency, again indicating that $\sim$2~kW is feasible on modest roof areas~\cite{Gagnon2016RooftopPVTechnicalPotential}.

Policy  design at the time also discouraged oversizing: California’s NSHP required systems to be ``sized so that the electricity produced offsets part or all of on-site needs,'' with a minimum of 1~kW AC and systems $\leq$7.5~kW AC deemed ``sized to serve on-site load.'' This aligns with our decision not to exceed historical maximum household loads in these datasets~\cite{CEC2013NSHPGuidebook}.

\subsubsection{Appliance State Labeling Thresholds}

The fixed thresholds for appliance state detection are based on typical operating powers observed in the datasets and are consistent with prior research~\cite{kelly2015neural,kolter2011redd}. Table~\ref{tab:appliance_thresholds} summarizes the thresholds used.

\begin{table}[h]
\centering
\caption{Appliance Power Thresholds for State Detection}
\label{tab:appliance_thresholds}
\begin{tabular}{lcc}
\toprule
\textbf{Appliance} & \textbf{Threshold (W)} & \textbf{Dataset}\\
\midrule
Microwave & 200 & REDD, UK-DALE\\
Fridge & 50 & REDD, UK-DALE\\
Dishwasher & 700 & REDD, UK-DALE\\
Washing Machine & 500 & REDD, UK-DALE\\
Kettle & 1500 & UK-DALE\\
\bottomrule
\end{tabular}
\end{table}

\subsubsection{Data Splitting}

The datasets are split into training and testing sets based on the dates specified in Section~\ref{subsec:simulated_injection}. 

\textbf{Dataset Configuration:} We evaluate our approach on two public datasets with the following setup:
\begin{itemize}
    \item \textbf{REDD~\cite{kolter2011redd}}: The dataset contains power sequence data for 6 US houses of 24 to 36 days between Apr. 2011 and Jun. 2011, with 1 Hz sampling frequency for the mains meter and 3 Hz for the 10 $\sim$ 25 types of appliance meters. We utilize Houses 1--3, with different monitoring periods for each house:
    \begin{itemize}
        \item House 1: Training period from 2011-04-19 to 2011-05-03, testing period from 2011-05-04 to 2011-05-11; monitored appliances include microwave, refrigerator, dishwasher, and washing machine.
        \item House 2: Training period from 2011-04-19 to 2011-04-25, testing period from 2011-04-26 to 2011-04-29; monitored appliances include microwave, refrigerator, and dishwasher.
        \item House 3: Training period from 2011-04-19 to 2011-04-25, testing period from 2011-04-26 to 2011-04-29; monitored appliances include microwave, refrigerator, dishwasher, and washing machine.
    \end{itemize}
    \item \textbf{UK-DALE~\cite{kelly2015uk}:} The dataset contains aggregate power consumption and measurements of 4$\sim$54 appliances from 5 UK houses. For house 1, the mains readings were recorded every 1 second, and appliances power readings were recorded every 6 seconds from Nov. 2012 to Jan. 2015, while for the other 4 houses, the recording periods are about half a year in 2013 or 2014. We analyze Houses 1--2, with consistent monitoring of five major appliances (kettle, microwave, refrigerator, dishwasher, and washing machine):
    \begin{itemize}
        \item House 1: Training period from 2013-04-01 to 2013-04-14, testing period from 2013-04-15 to 2013-04-21.
        \item House 2: Training period from 2013-06-01 to 2013-06-14, testing period from 2013-06-15 to 2013-06-21.
    \end{itemize}
\end{itemize}

For cross-validation purposes, we implement three distinct evaluation modes: (1) standard chronological split as specified above, (2) testing on the initial period while training on the remainder, and (3) testing on a mid-period segment while training on the surrounding data. For fair comparison, all test periods maintain consistent durations across different splits.

\subsubsection{Simulation Details}

Our data processing pipeline is implemented using Python and leverages libraries such as pandas for data manipulation and NumPy for numerical computations. 

\begin{table*}[h!]
\centering
\caption{Statistics of the Simulated Public Datasets (Based on Combined Training and Testing Data)} \label{tab:sim_dataset_stats_2kw}
\begin{tabular}{llccccc}
\hline
\textbf{Appliance} & \textbf{Threshold (W)} & \textbf{Max Power (W)} & \textbf{Mean Power (W)} & \textbf{Std Dev (W)} & \textbf{On-time (\%)} \\
\hline
\multicolumn{6}{c}{\textbf{REDD House 1 (Period: 2011-04-19 to 2011-05-11)}} \\
\hline
Aggregate Power     & -        & 5036.50   & 120.82   & 374.56   & -            \\
Micro-inverter      & -        & 1330.46   & 28.21    & 96.58    & 38.23\%      \\
Microwave           & 200.00   & 2905.00   & 18.71    & 146.26   & 1.00\%       \\
Fridge              & 50.00    & 2143.00   & 48.50    & 82.95    & 21.73\%      \\
Dish Washer         & 700.00   & 1396.00   & 21.00    & 137.61   & 1.52\%       \\
Washing Machine     & 500.00   & 3769.00   & 32.50    & 291.40   & 1.18\%       \\
\hline
\multicolumn{6}{c}{\textbf{REDD House 2 (Period: 2011-04-19 to 2011-04-29)}} \\
\hline
Aggregate Power     & -        & 2195.00   & 99.77    & 160.38   & -            \\
Micro-inverter      & -        & 1170.78   & 31.78    & 75.29    & 34.86\%      \\
Microwave           & 200.00   & 1967.00   & 14.43    & 100.72   & 0.32\%       \\
Fridge              & 50.00    & 2188.00   & 77.60    & 86.89    & 43.34\%      \\
Dish Washer         & 700.00   & 1457.00   & 7.74     & 89.20    & 0.54\%       \\
Washing Machine     & 500.00   & 28.00     & 2.04     & 0.49     & -\footnote{There is no washing machine in REDD house2 in reality.}\\
\hline
\multicolumn{6}{c}{\textbf{REDD House 3 (Period: 2011-04-19 to 2011-04-29)}} \\
\hline
Aggregate Power     & -        & 7036.00   & 126.08   & 569.50   & -            \\
Micro-inverter      & -        & 1330.46   & 15.89    & 51.70    & 11.29\%      \\
Microwave           & 200.00   & 1805.50   & 6.87     & 87.59    & 0.28\%       \\
Fridge              & 50.00    & 1550.00   & 42.32    & 63.74    & 33.90\%      \\
Dish Washer         & 200.00   & 780.00    & 5.35     & 54.65    & 0.72\%       \\
Washing Machine     & 500.00   & 5535.00   & 71.52    & 549.41   & 1.41\%       \\
\hline
\multicolumn{6}{c}{\textbf{UK-DALE House 1 (Period: 2013-04-01 to 2013-04-21)}} \\
\hline
Aggregate Power     & -        & 5432.00   & 118.77   & 391.46   & -            \\
Micro-inverter      & -        & 1289.01   & 33.00    & 118.07   & 18.92\%      \\
Kettle              & 1500.00  & 2761.00   & 18.17    & 199.54   & 0.73\%       \\
Microwave           & 200.00   & 3113.00   & 7.57     & 100.01   & 0.46\%       \\
Fridge              & 50.00    & 1902.00   & 35.70    & 52.15    & 38.32\%      \\
Dish Washer         & 700.00   & 2967.00   & 21.96    & 208.99   & 0.80\%       \\
Washing Machine     & 500.00   & 3918.00   & 35.14    & 231.69   & 1.52\%       \\
\hline
\multicolumn{6}{c}{\textbf{UK-DALE House 2 (Period: 2013-06-01 to 2013-06-21)}} \\
\hline
Aggregate Power     & -        & 6101.00   & 129.10   & 438.83   & -            \\
Micro-inverter      & -        & 1320.32   & 34.43    & 109.78   & 48.54\%      \\
Kettle              & 1500.00  & 3993.00   & 26.60    & 273.63   & 0.87\%       \\
Microwave           & 200.00   & 2500.00   & 7.63     & 98.18    & 0.58\%       \\
Fridge              & 50.00    & 1352.00   & 41.90    & 45.95    & 39.39\%      \\
Dish Washer         & 700.00   & 3955.00   & 42.62    & 282.46   & 2.04\%       \\
Washing Machine     & 500.00   & 2974.00   & 10.34    & 118.48   & 0.32\%       \\
\hline
\hline
\multicolumn{6}{l}{\textit{There is no washing machine in REDD house2 in reality. According to the meter readings of REDD house 2.}}
\end{tabular}
\end{table*}

\section{Details of Experiment}\label{appendix:experiment}

\subsection{Details of Benchmark Methods}\label{appendix:benchmarks}

To evaluate the performance of our proposed approach in the new scenario, we conducted benchmark experiments using several state-of-the-art models for appliance state identification and micro-inverter signal decomposition. All models were configured with carefully selected hyperparameters to ensure a fair comparison. The input data consisted of sequences of length 300, with two input features: active power and reactive power.

For \textbf{appliance state identification}, we compared the following methods:

\begin{itemize}
    \item \textbf{FHMM (Factorial Hidden Markov Model)}: A classical probabilistic model that represents the aggregate power consumption as the sum of multiple hidden Markov chains, each modeling an individual appliance's state~\cite{kim2011unsupervised}. We implemented FHMM using the \textit{hmmlearn} library, modeling each appliance with a two-state (ON/OFF) HMM. The model was trained using the Expectation-Maximization algorithm over sequences of length 300.

    \item \textbf{XGBoost}: An efficient implementation of gradient boosting decision trees capable of capturing complex nonlinear relationships in the data~\cite{chen2016xgboost}. We utilized the XGBoost classifier with parameters set as \texttt{use\_label\_encoder=False} and \texttt{eval\_metric='logloss'}. The model was trained on flattened input sequences, leveraging both active power and reactive power features.

    \item \textbf{Seq2Point}: A neural network architecture that maps a sequence of aggregate power readings to a point estimate of an individual appliance's state~\cite{zhang2018sequence}. Our implementation used a convolutional neural network with five convolutional layers:

    \begin{itemize}
        \item \textbf{Conv1}: 2 input channels, 30 filters, kernel size 10
        \item \textbf{Conv2}: 30 filters, kernel size 8
        \item \textbf{Conv3}: 40 filters, kernel size 6
        \item \textbf{Conv4}: 50 filters, kernel size 5
        \item \textbf{Conv5}: 50 filters, kernel size 5
    \end{itemize}

    Each convolutional layer was followed by a ReLU activation function. After flattening, a fully connected layer with 1,024 units and ReLU activation was applied, followed by a dropout layer with a rate of 0.2. The output layer had units equal to the number of appliance classes, using a sigmoid activation function for binary classification.

    \item \textbf{CNN-LSTM}: A hybrid deep learning model combining Convolutional Neural Networks (CNNs) for feature extraction and Long Short-Term Memory (LSTM) networks for temporal sequence modeling~\cite{kaselimi2020multi}. The CNN component consisted of:

    \begin{itemize}
        \item \textbf{Conv1}: 64 filters, kernel size 3, padding 1
        \item \textbf{Conv2}: 64 filters, kernel size 3, padding 1
    \end{itemize}

    Each convolutional layer was followed by a ReLU activation and a max-pooling layer. The output was reshaped and passed to an LSTM layer with 128 hidden units. A fully connected layer with sigmoid activation provided the final appliance state predictions.

    \item \textbf{Transformer}: A deep learning model based on the self-attention mechanism, allowing for efficient sequence modeling without relying on recurrence~\cite{vaswani2017attention}. We implemented a Transformer encoder with the following parameters:

    \begin{itemize}
        \item \textbf{Input Embedding}: Linear layer mapping 2 input features to a model dimension of 128
        \item \textbf{Positional Encoding}: Added to input embeddings to retain temporal information
        \item \textbf{Encoder Layers}: 2 layers, each with 4 attention heads and a feedforward network dimension of 256
        \item \textbf{Dropout Rate}: 0.1
    \end{itemize}

    The output from the final time step was passed through a linear layer to map to the number of appliance classes, followed by a sigmoid activation function.

    \item \textbf{UNetNILM}: An adaptation of the U-Net architecture for NILM tasks, designed to capture multiscale temporal patterns in the data~\cite{faustine2020unet}. The model featured an encoder-decoder structure with skip connections:

    \textbf{Encoder}:

    \begin{itemize}
        \item \textbf{Conv1}: 2 input channels, 16 filters, kernel size 3, padding 1
        \item \textbf{Conv2}: 16 filters, 32 filters, kernel size 3, padding 1
        \item \textbf{Conv3}: 32 filters, 64 filters, kernel size 3, padding 1
        \item \textbf{Conv4}: 64 filters, 128 filters, kernel size 3, padding 1
    \end{itemize}

    Max-pooling layers with kernel size 2 and stride 2 were applied after each convolutional layer.

    \textbf{Decoder}:

    \begin{itemize}
        \item \textbf{UpConv1}: Transposed convolution from 128 to 64 filters, kernel size 2, stride 2, output padding 1
        \item \textbf{DecConv1}: Convolutional layer after concatenation, 128 input filters, 64 output filters
        \item \textbf{UpConv2}: Transposed convolution from 64 to 32 filters, kernel size 2, stride 2
        \item \textbf{DecConv2}: Convolutional layer after concatenation, 64 input filters, 32 output filters
        \item \textbf{UpConv3}: Transposed convolution from 32 to 16 filters, kernel size 2, stride 2
        \item \textbf{DecConv3}: Convolutional layer after concatenation, 32 input filters, 16 output filters
    \end{itemize}

    The model had two output heads:

    \begin{itemize}
        \item \textbf{Classification Head}: Convolutional layer mapping 16 filters to 7 appliance classes, followed by a sigmoid activation. The classification output was taken from the last time step.
        \item \textbf{Regression Head}: Convolutional layer mapping 16 filters to 1 output, providing the estimated power consumption sequence.
    \end{itemize}

    The model was trained to jointly predict appliance states and estimate power consumption sequences, leveraging the encoder-decoder structure to capture both local and global features.

\end{itemize}

For the \textbf{micro-inverter signal decomposition}, we employed the following benchmark methods:

\begin{itemize}
    \item \textbf{FHMM}: Similar to its application in appliance state identification, FHMM was used to model the aggregate power consumption, focusing on isolating the micro-inverter signal. Each appliance, including the micro-inverter, was modeled as a separate HMM.

    \item \textbf{Seq2Seq}: A sequence-to-sequence neural network model that learns to map input sequences of aggregate power to output sequences of disaggregated signals~\cite{du2016regression}. Our implementation used an encoder-decoder LSTM architecture with:

    \begin{itemize}
        \item \textbf{Input Size}: 2 features (active power and reactive power)
        \item \textbf{Hidden Size}: 64 units
        \item \textbf{Number of Layers}: 2
    \end{itemize}

    The encoder processed the input sequence, and the decoder generated the estimated micro-inverter power sequence. A fully connected layer mapped the decoder outputs to the final predictions.

    \item \textbf{DAE (Denoising Autoencoder)}: An autoencoder model trained to reconstruct the target signal from a corrupted version of the input, effectively isolating the micro-inverter's contribution from the aggregate power~\cite{jia2019matrix}. The model consisted of:

    \textbf{Encoder}:

    \begin{itemize}
        \item \textbf{Input Layer}: Flattened sequence of length 300
        \item \textbf{Linear Layer 1}: 300 inputs to 128 units, ReLU activation
        \item \textbf{Linear Layer 2}: 128 units to 64 units, ReLU activation
    \end{itemize}

    \textbf{Decoder}:

    \begin{itemize}
        \item \textbf{Linear Layer 3}: 64 units to 128 units, ReLU activation
        \item \textbf{Output Layer}: 128 units to 300 outputs
    \end{itemize}

    The output was reshaped back to the sequence format, representing the reconstructed micro-inverter signal.

    \item \textbf{BERT+}~\cite{li2024bert} baseline following Li et al. (2024) architecture description on PV injection estimation task. We use  5-layer BERT blocks, a feed-forward dimension of 180, and positional encoding. Since BERT+ requires external weather data for PV estimation, we use: (i) Active power (P) from aggregated consumption with injection, (ii) Reactive power (Q), (iii) Time positional embeddings (hour\_sin, hour\_cos), and (iv) Weather features from NSRDB (GHI and Temperature).
    
    \item \textbf{SunDance}~\cite{chen2017sundance} for PV injection estimation; this method was originally designed for solar-only disaggregation from net meter data without appliance monitoring. It is a ``black-box'' method combining physical solar modeling with machine learning. We first construct a location-specific clear sky generation model using the geographical coordinates and NSRDB solar irradiance data. This establishes the theoretical maximum solar generation potential at each timestamp based on solar position and panel characteristics. Following the paper~\cite{chen2017sundance}, which identified a universal weather-solar effect, we train an SVM with an RBF kernel to map weather metrics to the fraction of maximum clear sky generation. The estimated PV is then subtracted from the aggregate signal. We use SVM-RBF kernel (C=100, gamma='scale') settings.
    
    \item \textbf{UNetNILM}: Utilizing the same architecture as in the appliance state identification task, UNetNILM was applied to the micro-inverter signal decomposition. The regression output provided the estimated power sequence of the micro-inverter. The model effectively captured both global and local temporal patterns through its encoder-decoder structure with skip connections.

\end{itemize}

These benchmark methods offer a comprehensive evaluation of different modeling approaches for NILM tasks. FHMM serves as a foundational baseline, allowing us to compare traditional probabilistic methods with modern machine-learning techniques. XGBoost provides a powerful gradient-boosting framework for classification tasks. Deep learning models like Seq2Point, CNN-LSTM, Transformer, and UNetNILM exploit advanced architectures to capture intricate patterns in the data. The Transformer model leverages self-attention mechanisms for improved performance, while UNetNILM benefits from multiscale feature learning through its encoder-decoder architecture with skip connections.

For the micro-inverter decomposition task, Seq2Seq models effectively handle sequence mapping, DAE focuses on learning robust representations to reconstruct the micro-inverter signals, and UNetNILM offers joint modeling of classification and regression tasks, enhancing disaggregation performance.

All models were trained using the Adam optimizer, with learning rates and training epochs carefully adjusted for each model to ensure optimal convergence. The models were evaluated based on their ability to accurately identify appliance states and decompose the micro-inverter signal from the aggregated power data.

\begin{figure*}[h]
    \centering
    \includegraphics[width=0.85\linewidth]{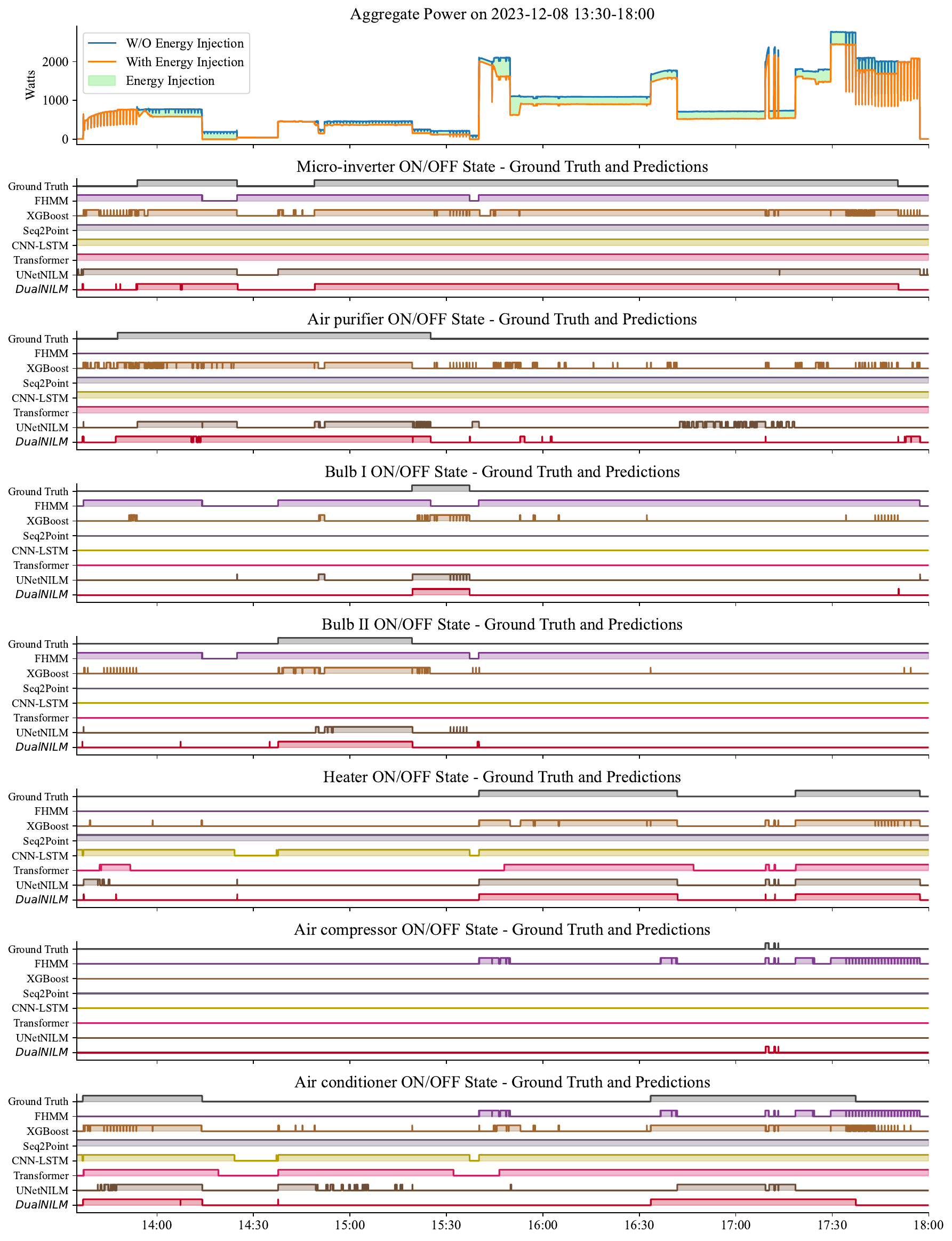}
    \caption{Visualization of the Predictions of Appliances ON/OFF State on Dec.8, 2023 of our dataset}
    \label{fig:1208model_comparison_fullappendix}
\end{figure*}

\section{Full Experiment Results}\label{appendix:more_experiment_result_2kw}

In this section, we will present the full version of all experiment results on all datasets, including our laboratory dataset and synthetic datasets.

\subsection{Overview of All Results Cross All Datasets}
Table~\ref{tab:total_overall_avg_appendix_2kw} and  \ref{tab:total_overall_digg_avg} show the overall average performance metrics on appliance state recognition and energy injection disaggregation tasks.

\begin{table*}[h]
\centering
\caption{Overall Avg. Performance Metrics (Cross-validation on all splits of each Dataset)}\label{tab:total_overall_avg_appendix_2kw}
\vspace{-5pt}
\resizebox{0.85\textwidth}{!}{
\setlength{\tabcolsep}{1.2mm}
\renewcommand{\arraystretch}{0.97}
\newcommand{\meanstd}[2]{#1$_{\scriptstyle\,\pm#2}$}
\begin{tabular}{c|c|c|cccc}
\hline
\multicolumn{2}{c|}{\makecell[c]{Dataset}} & \makecell[c]{Method} & Accuracy~(\%) & Recall~(\%) & Precision~(\%) & F1~(\%) \\
\hline
\multicolumn{2}{c|}{\multirow{7}{*}{\centering \makecell[c]{Laboratory Dataset \\ (Real BTM Injection)}}} & FHMM & \meanstd{60.95}{27.96} & \meanstd{28.32}{40.81} & \meanstd{16.84}{28.20} & \meanstd{17.25}{27.23} \\
\multicolumn{2}{c|}{} & XGBoost & \meanstd{85.82}{2.07} & \meanstd{48.20}{7.69} & \meanstd{52.10}{6.02} & \meanstd{47.75}{6.75} \\
\multicolumn{2}{c|}{} & Seq2Point & \meanstd{67.63}{17.08} & \meanstd{53.06}{10.80} & \meanstd{27.86}{14.16} & \meanstd{34.27}{13.29} \\
\multicolumn{2}{c|}{} & CNN-LSTM & \meanstd{74.03}{15.51} & \meanstd{52.97}{11.02} & \meanstd{30.29}{14.28} & \meanstd{36.67}{13.25} \\
\multicolumn{2}{c|}{} & Transformer & \meanstd{81.79}{12.65} & \meanstd{52.91}{10.76} & \meanstd{38.16}{15.53} & \meanstd{42.76}{14.02} \\  
\multicolumn{2}{c|}{} & UNetNILM & \meanstd{89.20}{9.76} & \meanstd{54.01}{27.35} & \meanstd{60.81}{30.96} & \meanstd{53.89}{27.00} \\
\multicolumn{2}{c|}{} & \textbf{\ouralg} & \textbf{\meanstd{98.54}{1.71}} & \textbf{\meanstd{85.43}{28.03}} & \textbf{\meanstd{84.16}{28.45}} & \textbf{\meanstd{84.54}{27.93}} \\
\hline
\multirow{21}{*}{\shortstack{REDD\\ (Synthetic PV Injection)}}
& \multirow{7}{*}{House 1} & FHMM & \meanstd{79.52}{39.21} & \meanstd{1.46}{3.60} & \meanstd{13.47}{33.46} & \meanstd{2.07}{6.20} \\
& & XGBoost & \meanstd{97.90}{0.38} & \meanstd{74.50}{2.64} & \meanstd{88.74}{2.08} & \meanstd{79.18}{1.79} \\
& & Seq2Point & \meanstd{97.65}{0.46} & \meanstd{42.20}{3.43} & \meanstd{56.42}{1.16} & \meanstd{46.07}{4.03} \\
& & CNN-LSTM & \meanstd{98.67}{0.46} & \meanstd{46.32}{2.41} & \meanstd{50.75}{3.19} & \meanstd{47.96}{2.07} \\
& & Transformer & \meanstd{95.99}{1.10} & \meanstd{56.08}{5.37} & \meanstd{54.10}{8.04} & \meanstd{53.42}{6.31} \\
& & UNetNILM &  \meanstd{99.05}{0.31} & \meanstd{76.11}{3.42} & \meanstd{85.22}{5.99} & \meanstd{75.27}{4.25} \\
& & \textbf{\ouralg} & \textbf{\meanstd{99.84}{0.07}} & \textbf{\meanstd{96.26}{3.15}} & \textbf{\meanstd{97.11}{1.90}} & \textbf{\meanstd{96.64}{2.42}} \\
\cline{2-7}
& \multirow{7}{*}{House 2}& FHMM        & \meanstd{75.08}{42.51}        & \meanstd{0.58}{1.35}          & \meanstd{0.34}{0.79}          & \meanstd{0.43}{1.00}          \\
&  & XGBoost     & \meanstd{98.43}{0.09}         & \meanstd{92.47}{3.94}         & \textbf{\meanstd{95.58}{1.21}} & \textbf{\meanstd{93.53}{3.15}} \\
&  & Seq2Point   & \meanstd{91.43}{1.50}         & \meanstd{52.90}{2.47}         & \meanstd{58.44}{1.59}         & \meanstd{50.85}{2.27}         \\
&  & CNN-LSTM    & \meanstd{98.86}{0.32}         & \meanstd{49.19}{0.39}         & \meanstd{48.74}{0.15}         & \meanstd{48.96}{0.26}         \\
&  & Transformer & \meanstd{97.92}{0.19}         & \meanstd{47.78}{0.35}         & \meanstd{47.98}{0.50}         & \meanstd{47.87}{0.22}         \\
&  & UNetNILM    & \meanstd{99.57}{0.22}         & \meanstd{64.16}{16.03}        & \meanstd{77.22}{20.61}        & \meanstd{67.28}{17.02}        \\
&  & \textbf{\ouralg}   & \textbf{\meanstd{99.70}{0.22}} & \textbf{\meanstd{92.48}{5.55}} & \meanstd{94.00}{5.47}         & \meanstd{93.11}{4.96}         \\
\cline{2-7}
& \multirow{7}{*}{House 3} & FHMM        & \meanstd{83.84}{27.80}          & \meanstd{4.53}{8.79}          & \meanstd{0.90}{1.69}           & \meanstd{1.49}{2.80}           \\
& & XGBoost     & \meanstd{94.89}{2.15}          & \meanstd{46.22}{0.48}         & \meanstd{56.38}{2.49}          & \meanstd{50.51}{1.28}          \\
& & Seq2Point   & \meanstd{98.10}{0.88}          & \meanstd{42.78}{7.36}         & \meanstd{50.90}{11.98}         & \meanstd{44.47}{7.97}          \\
& & CNN-LSTM    & \meanstd{95.92}{2.70}          & \meanstd{28.85}{5.34}         & \meanstd{38.17}{0.75}          & \meanstd{32.62}{3.48}          \\
& & Transformer & \meanstd{97.81}{0.61}          & \meanstd{40.21}{5.32}         & \meanstd{38.89}{4.60}          & \meanstd{39.43}{4.67}          \\
& & UNetNILM    & \meanstd{99.07}{0.32}          & \meanstd{59.91}{0.04}         & \meanstd{54.73}{1.56}          & \meanstd{57.10}{0.95}          \\
& & \textbf{\ouralg}    & \textbf{\meanstd{99.69}{0.24}} & \textbf{\meanstd{77.42}{17.13}}& \textbf{\meanstd{84.46}{12.18}} & \textbf{\meanstd{80.04}{15.51}} \\
\hline
\multirow{14}{*}{\shortstack{UKDALE\\ (Synthetic PV Injection)}}& \multirow{7}{*}{House 1} & FHMM         & \meanstd{89.25}{23.05}          & \meanstd{1.58}{3.57}           & \meanstd{0.76}{1.87}            & \meanstd{1.01}{2.39}            \\
& & XGBoost      & \meanstd{93.64}{0.42}           & \meanstd{41.74}{1.40}          & \meanstd{68.04}{3.98}           & \meanstd{45.97}{2.56}           \\
& & Seq2Point    & \meanstd{88.70}{0.58}           & \meanstd{41.47}{2.83}          & \meanstd{43.29}{5.22}           & \meanstd{36.79}{3.32}           \\
& & CNN-LSTM     & \meanstd{98.33}{0.24}           & \meanstd{31.76}{0.20}          & \meanstd{31.29}{0.24}           & \meanstd{31.52}{0.15}           \\
& & Transformer  & \meanstd{96.90}{0.29}           & \meanstd{33.65}{1.03}          & \meanstd{27.76}{2.49}           & \meanstd{30.26}{1.69}           \\
& & UNetNILM     & \meanstd{98.76}{0.16}           & \meanstd{52.03}{5.14}          & \meanstd{72.01}{18.92}          & \meanstd{55.20}{7.02}           \\
& & \textbf{\ouralg}
                & \textbf{\meanstd{99.72}{0.21}}  & \textbf{\meanstd{95.23}{3.53}} & \textbf{\meanstd{94.30}{4.09}}  & \textbf{\meanstd{94.71}{3.63}}  \\
\cline{2-7}
& \multirow{7}{*}{House 2} & FHMM         & \meanstd{82.95}{36.49}          & \meanstd{0.62}{1.17}           & \meanstd{0.52}{1.01}            & \meanstd{0.55}{1.06}            \\
& & XGBoost      & \meanstd{96.13}{0.61}          & \meanstd{66.61}{5.32}          & \meanstd{79.19}{9.68}           & \meanstd{70.67}{6.81}           \\
& & Seq2Point    & \meanstd{98.83}{0.18}          & \meanstd{42.68}{3.09}          & \meanstd{46.49}{12.53}          & \meanstd{41.92}{3.19}           \\
& & CNN-LSTM     & \meanstd{97.81}{0.86}          & \meanstd{41.65}{5.05}          & \meanstd{45.48}{3.20}           & \meanstd{43.33}{4.21}           \\
& & Transformer  & \meanstd{99.13}{0.14}          & \meanstd{59.88}{5.10}          & \meanstd{77.44}{7.86}           & \meanstd{61.73}{4.66}           \\
& & UNetNILM     & \meanstd{99.17}{0.19}          & \meanstd{64.25}{5.53}          & \meanstd{68.74}{5.74}           & \meanstd{65.43}{4.72}           \\
& & \textbf{\ouralg}
                & \textbf{\meanstd{99.73}{0.04}}  & \textbf{\meanstd{90.63}{5.01}} & \textbf{\meanstd{94.46}{3.36}}  & \textbf{\meanstd{92.31}{3.76}}  \\
\bottomrule
\end{tabular}}
\end{table*}

\begin{table*}[htp]
\centering
\caption{Overall Avg. Performance Metrics of Energy Injection Disaggregation (Cross-validation on all splits of each Dataset)}\label{tab:total_overall_digg_avg}
\vspace{-2pt}
\resizebox{0.55\textwidth}{!}{
\setlength{\tabcolsep}{1.2mm}
\renewcommand{\arraystretch}{0.97}
\newcommand{\meanstd}[2]{#1$_{\scriptstyle\,\pm#2}$}
\begin{tabular}{c|c|c|cccc}
\hline
\multicolumn{2}{c|}{\makecell[c]{Dataset}} & \makecell[c]{Method} & \multicolumn{2}{c}{\makecell[c]{Avg. RMSE~$\downarrow$}} & \multicolumn{2}{c}{\makecell[c]{Avg. MAE~$\downarrow$}} \\
\hline
\multicolumn{2}{c|}{\multirow{5}{*}{\centering \makecell[c]{Laboratory Dataset \\ (Real Injection)}}} & FHMM &  \multicolumn{2}{c}{\meanstd{0.5015}{0.0823}}& \multicolumn{2}{c}{\meanstd{0.4492}{0.0867}} \\
\multicolumn{2}{c|}{} & Seq2Seq &  \multicolumn{2}{c}{\meanstd{0.3512}{0.1111}}& \multicolumn{2}{c}{\meanstd{0.2622}{0.0888}} \\
\multicolumn{2}{c|}{} & DAE &   \multicolumn{2}{c}{ \meanstd{0.8830}{0.0803}}& \multicolumn{2}{c}{ \meanstd{0.7851}{0.1410}} \\
\multicolumn{2}{c|}{} & UNetNILM & \multicolumn{2}{c}{ \meanstd{0.2305}{0.0858}} & \multicolumn{2}{c}{ \meanstd{0.1883}{0.0802}} \\
\multicolumn{2}{c|}{} & \textbf{\ouralg} & \multicolumn{2}{c}{\textbf{ \meanstd{0.1429}{0.0896}}} & \multicolumn{2}{c}{\textbf{\meanstd{0.0804}{0.0677}}} \\
\hline
\multirow{28}{*}{\shortstack{REDD\\with\\Synthetic\\Injection}}
& \multirow{7}{*}{House 1} & FHMM      & \multicolumn{2}{c}{ \meanstd{0.8093}{0.0141}} & \multicolumn{2}{c}{ \meanstd{0.2200}{0.0054}} \\
&                          & SunDance & \multicolumn{2}{c}{\meanstd{0.8523}{0.0892}} & \multicolumn{2}{c}{\meanstd{0.2847}{0.0421}} \\
&                          & Seq2Seq   & \multicolumn{2}{c}{\textbf{ \meanstd{0.6834}{0.0217}}} & \multicolumn{2}{c}{\textbf{ \meanstd{0.2370}{0.0121}}} \\
&                          & DAE       & \multicolumn{2}{c}{ \meanstd{0.8115}{0.0172}} & \multicolumn{2}{c}{ \meanstd{0.2171}{0.0069}} \\
&                          & BERT+    & \multicolumn{2}{c}{\meanstd{0.7234}{0.0167}} & \multicolumn{2}{c}{\meanstd{0.1945}{0.0089}} \\
&                          & UNetNILM  & \multicolumn{2}{c}{ \meanstd{0.7589}{0.1951}} & \multicolumn{2}{c}{ \meanstd{0.1983}{0.0164}} \\
&                          & \textbf{\ouralg} & \multicolumn{2}{c}{ \meanstd{0.6661}{0.0145}} & \multicolumn{2}{c}{ \meanstd{0.1777}{0.0056}} \\
\cline{2-7}
& \multirow{7}{*}{House 2} & FHMM      & \multicolumn{2}{c}{ \meanstd{0.8403}{0.1384}} & \multicolumn{2}{c}{ \meanstd{0.3367}{0.0629}} \\
&                          & SunDance & \multicolumn{2}{c}{\meanstd{0.9271}{0.1687}} & \multicolumn{2}{c}{\meanstd{0.3542}{0.0823}} \\
&                          & Seq2Seq   & \multicolumn{2}{c}{ \meanstd{0.7217}{0.1521}} & \multicolumn{2}{c}{ \meanstd{0.3287}{0.0624}} \\
&                          & DAE       & \multicolumn{2}{c}{ \meanstd{0.8448}{0.1703}} & \multicolumn{2}{c}{ \meanstd{0.3356}{0.0782}} \\
&                          & BERT+    & \multicolumn{2}{c}{\meanstd{0.7812}{0.1534}} & \multicolumn{2}{c}{\meanstd{0.3012}{0.0701}} \\
&                          & UNetNILM  & \multicolumn{2}{c}{ \meanstd{2.2077}{0.6670}} & \multicolumn{2}{c}{ \meanstd{0.6601}{0.1903}} \\
&                          & \textbf{\ouralg} & \multicolumn{2}{c}{ \textbf{ \meanstd{0.7065}{0.1419}}} & \multicolumn{2}{c}{ \textbf{ \meanstd{0.2768}{0.0648}}} \\
\cline{2-7}
& \multirow{7}{*}{House 3} & FHMM      & \multicolumn{2}{c}{ \meanstd{0.5575}{0.0574}} & \multicolumn{2}{c}{ \meanstd{0.1766}{0.0532}} \\
&                          & SunDance & \multicolumn{2}{c}{\meanstd{0.5678}{0.0834}} & \multicolumn{2}{c}{\meanstd{0.2134}{0.0712}} \\
&                          & Seq2Seq   & \multicolumn{2}{c}{ \meanstd{0.4974}{0.0543}} & \multicolumn{2}{c}{ \meanstd{0.1940}{0.0505}} \\
&                          & DAE       & \multicolumn{2}{c}{ \meanstd{0.5606}{0.0704}} & \multicolumn{2}{c}{ \meanstd{0.1743}{0.0660}} \\
&                          & BERT+    & \multicolumn{2}{c}{\meanstd{0.4723}{0.0645}} & \multicolumn{2}{c}{\meanstd{0.1623}{0.0598}} \\
&                          & UNetNILM  & \multicolumn{2}{c}{ \meanstd{2.2517}{0.2121}} & \multicolumn{2}{c}{ \meanstd{0.4708}{0.1640}} \\
&                          & \textbf{\ouralg} & \multicolumn{2}{c}{ \textbf{ \meanstd{0.4261}{0.0582}}} & \multicolumn{2}{c}{ \textbf{ \meanstd{0.1422}{0.0546}}} \\
\cline{2-7}
& \multirow{7}{*}{Avg.}    & FHMM      & \multicolumn{2}{c}{ \meanstd{0.7357}{0.0700}} & \multicolumn{2}{c}{ \meanstd{0.2444}{0.0405}} \\
&                          & SunDance & \multicolumn{2}{c}{\meanstd{0.7824}{0.1138}} & \multicolumn{2}{c}{\meanstd{0.2841}{0.0652}} \\
&                          & Seq2Seq   & \multicolumn{2}{c}{ \meanstd{0.6342}{0.0760}} & \multicolumn{2}{c}{ \meanstd{0.2532}{0.0417}} \\
&                          & DAE       & \multicolumn{2}{c}{ \meanstd{0.7390}{0.0860}} & \multicolumn{2}{c}{ \meanstd{0.2423}{0.0504}} \\
&                          & BERT+    & \multicolumn{2}{c}{\meanstd{0.6590}{0.0782}} & \multicolumn{2}{c}{\meanstd{0.2193}{0.0463}} \\
&                          & UNetNILM  & \multicolumn{2}{c}{ \meanstd{1.7394}{0.3581}} & \multicolumn{2}{c}{ \meanstd{0.4431}{0.1236}} \\
&                          & \textbf{\ouralg} & \multicolumn{2}{c}{ \textbf{ \meanstd{0.5996}{0.0715}}} & \multicolumn{2}{c}{ \textbf{ \meanstd{0.1989}{0.0417}}} \\
\hline
\multirow{21}{*}{\shortstack{UKDALE\\with\\Synthetic\\Injection}}
& \multirow{7}{*}{House 1} & FHMM      & \multicolumn{2}{c}{ \meanstd{1.1905}{0.0936}} & \multicolumn{2}{c}{ \meanstd{0.3331}{0.0015}} \\
&                          & SunDance & \multicolumn{2}{c}{\meanstd{1.1523}{0.1245}} & \multicolumn{2}{c}{\meanstd{0.3478}{0.0234}} \\
&                          & Seq2Seq   & \multicolumn{2}{c}{ \meanstd{0.9055}{0.0211}} & \multicolumn{2}{c}{ \meanstd{0.2758}{0.0103}} \\
&                          & DAE       & \multicolumn{2}{c}{ \meanstd{0.9197}{0.0255}} & \multicolumn{2}{c}{ \meanstd{0.2948}{0.0128}} \\
&                          & BERT+    & \multicolumn{2}{c}{\meanstd{0.9712}{0.1034}} & \multicolumn{2}{c}{\meanstd{0.2945}{0.0089}} \\
&                          & UNetNILM  & \multicolumn{2}{c}{ \meanstd{7.5014}{0.5703}} & \multicolumn{2}{c}{ \meanstd{1.3871}{0.1633}} \\
&                          & \textbf{\ouralg} & \multicolumn{2}{c}{\textbf{ \meanstd{0.9008}{0.0957}}} & \multicolumn{2}{c}{\textbf{ \meanstd{0.2731}{0.0020}}} \\
\cline{2-7}
& \multirow{7}{*}{House 2} & FHMM      & \multicolumn{2}{c}{ \meanstd{1.1477}{0.0213}} & \multicolumn{2}{c}{ \meanstd{0.3521}{0.0101}} \\
&                          & SunDance & \multicolumn{2}{c}{\meanstd{1.2187}{0.0534}} & \multicolumn{2}{c}{\meanstd{0.3687}{0.0178}} \\
&                          & Seq2Seq   & \multicolumn{2}{c}{ \textbf{ \meanstd{0.9252}{0.0256}}} & \multicolumn{2}{c}{ \textbf{ \meanstd{0.2833}{0.0165}}} \\
&                          & DAE       & \multicolumn{2}{c}{ \meanstd{0.9594}{0.0309}} & \multicolumn{2}{c}{ \meanstd{0.3368}{0.0202}} \\
&                          & BERT+    & \multicolumn{2}{c}{\meanstd{1.0123}{0.0287}} & \multicolumn{2}{c}{\meanstd{0.3123}{0.0134}} \\
&                          & UNetNILM  & \multicolumn{2}{c}{ \meanstd{1.1285}{0.0022}} & \multicolumn{2}{c}{ \meanstd{0.3043}{0.0046}} \\
&                          & \textbf{\ouralg} & \multicolumn{2}{c}{ \textbf{ \meanstd{0.9344}{0.0216}}} & \multicolumn{2}{c}{ \textbf{ \meanstd{0.2883}{0.0103}}} \\
\cline{2-7}
& \multirow{7}{*}{Avg.}    & FHMM      & \multicolumn{2}{c}{ \meanstd{1.1691}{0.0575}} & \multicolumn{2}{c}{ \meanstd{0.3426}{0.0058}} \\
&                          & SunDance & \multicolumn{2}{c}{\meanstd{1.1855}{0.0890}} & \multicolumn{2}{c}{\meanstd{0.3583}{0.0206}} \\
&                          & Seq2Seq   & \multicolumn{2}{c}{ \textbf{ \meanstd{0.9153}{0.0234}}} & \multicolumn{2}{c}{ \textbf{ \meanstd{0.2795}{0.0134}}} \\
&                          & DAE       & \multicolumn{2}{c}{ \meanstd{0.9395}{0.0282}} & \multicolumn{2}{c}{ \meanstd{0.3158}{0.0165}} \\
&                          & BERT+    & \multicolumn{2}{c}{\meanstd{0.9917}{0.0660}} & \multicolumn{2}{c}{\meanstd{0.3034}{0.0112}} \\
&                          & UNetNILM  & \multicolumn{2}{c}{ \meanstd{4.3150}{0.2863}} & \multicolumn{2}{c}{ \meanstd{0.8457}{0.0839}} \\
&                          & \textbf{\ouralg} & \multicolumn{2}{c}{ \textbf{ \meanstd{0.9176}{0.0586}}} & \multicolumn{2}{c}{ \textbf{ \meanstd{0.2807}{0.0062}}} \\
\hline
\end{tabular}}
\end{table*}

\subsection{Laboratory Dataset Results}

Table~\ref{tab:lab_results_full} shows the details of the laboratory dataset performance metrics for \ouralg and benchmarks.

Figure~\ref{fig:1208model_comparison_fullappendix} shows all seven different appliances (including micro-inverter)'s ground truth state, \ouralg, and benchmarks prediction on Dec.8, 2023 of our dataset.

\begin{table*}[h]
\centering
\caption{Laboratory Dataset Performance Metrics (cross-validation on all dates)}\label{tab:lab_results_full}
\vspace{-5pt}
\resizebox{0.70\textwidth}{!}{
\setlength{\tabcolsep}{1.2mm}
\renewcommand{\arraystretch}{0.97}
\newcommand{\meanstd}[2]{#1$_{\scriptstyle\,\pm#2}$}
\begin{tabular}{c|c|cccc}
\hline
\makecell[c]{Appliance} & \makecell[c]{Method} & Accuracy~(\%) & Recall~(\%) & Precision~(\%) & F1~(\%) \\
\hline
\multirow{7}{*}{Air compressor} 
& FHMM & \meanstd{99.56}{0.20} & \meanstd{14.29}{34.99} & \meanstd{14.29}{34.99} & \meanstd{14.29}{34.99}\\
& XGBoost & \meanstd{99.55}{0.23} & \meanstd{0.00}{0.00} & \meanstd{0.00}{0.00} & \meanstd{0.00}{0.00} \\
& Seq2Point & \meanstd{99.52}{0.25} & \meanstd{0.00}{0.00} & \meanstd{0.00}{0.00} & \meanstd{0.00}{0.00} \\
& CNN-LSTM & \meanstd{99.52}{0.25} & \meanstd{0.00}{0.00} & \meanstd{0.00}{0.00} & \meanstd{0.00}{0.00} \\
& Transformer & \meanstd{99.52}{0.25} & \meanstd{0.00}{0.00} & \meanstd{0.00}{0.00} & \meanstd{0.00}{0.00} \\
& UNetNILM & \meanstd{99.52}{0.25} & \meanstd{0.00}{0.00} & \meanstd{0.00}{0.00} & \meanstd{0.00}{0.00} \\
& \textbf{\ouralg} & \textbf{\meanstd{99.97}{0.03}} & \textbf{\meanstd{83.77}{37.10}} & \textbf{\meanstd{83.40}{36.97}} & \textbf{\meanstd{83.52}{36.91}} \\
\hline
\end{tabular}}
\end{table*}

\subsection{REDD Dataset Results}

\subsubsection{Full Results on House 1} See Table~\ref{tab:redd_house1_results_2kw}.
\begin{table*}[h]
\centering
\caption{REDD House 1 Performance Metrics (cross-validation on all splits)}\label{tab:redd_house1_results_2kw}
\vspace{-5pt}
\resizebox{0.70\textwidth}{!}{
\setlength{\tabcolsep}{1.2mm}
\renewcommand{\arraystretch}{0.97}
\newcommand{\meanstd}[2]{#1$_{\scriptstyle\,\pm#2}$}
\begin{tabular}{c|c|cccc}
\hline
\makecell[c]{Appliance} & \makecell[c]{Method} & Accuracy~(\%) & Recall~(\%) & Precision~(\%) & F1~(\%) \\
\hline
\multirow{7}{*}{Microwave} & FHMM & \meanstd{99.13}{0.53} & \meanstd{0.06}{0.09} & \meanstd{33.33}{47.14} & \meanstd{0.12}{0.17} \\
 & XGBoost & \meanstd{99.40}{0.32} & \meanstd{44.51}{21.18} & \meanstd{72.75}{8.68} & \meanstd{52.40}{17.83} \\
 & Seq2Point & \meanstd{99.37}{0.60} & \meanstd{35.85}{15.53} & \textbf{\meanstd{98.67}{0.41}} & \meanstd{51.25}{17.22} \\
 & CNN-LSTM & \meanstd{99.11}{0.66} & \meanstd{0.00}{0.00} & \meanstd{0.00}{0.00} & \meanstd{0.00}{0.00} \\
 & Transformer & \meanstd{99.11}{0.66} & \meanstd{0.00}{0.00} & \meanstd{0.00}{0.00} & \meanstd{0.00}{0.00} \\
 & UNetNILM & \meanstd{99.22}{0.63} & \meanstd{20.76}{10.51} & \meanstd{84.45}{15.52} & \meanstd{31.96}{14.16} \\
 & \textbf{\ouralg} & \textbf{\meanstd{99.90}{0.07}} & \textbf{\meanstd{88.10}{10.07}} & \meanstd{95.87}{5.23} & \textbf{\meanstd{91.75}{7.75}} \\
\bottomrule
\end{tabular}}
\end{table*}

\subsubsection{Full Results on House 2} See Table~\ref{tab:redd_house2_results_2kw}.
\begin{table*}[h]
\centering
\caption{REDD House 2 Performance Metrics (cross-validation on all splits)}\label{tab:redd_house2_results_2kw}
\vspace{-5pt}
\resizebox{0.70\textwidth}{!}{
\setlength{\tabcolsep}{1.2mm}
\renewcommand{\arraystretch}{0.97}
\newcommand{\meanstd}[2]{#1$_{\scriptstyle\,\pm#2}$}
\begin{tabular}{c|c|cccc}
\hline
\makecell[c]{Appliance} & \makecell[c]{Method} & Accuracy~(\%) & Recall~(\%) & Precision~(\%) & F1~(\%) \\
\hline
\multirow{7}{*}{Microwave}
 & FHMM        & \meanstd{99.56}{0.21}         & \meanstd{0.00}{0.00}          & \meanstd{0.00}{0.00}          & \meanstd{0.00}{0.00}          \\
 & XGBoost   & \textbf{\meanstd{99.93}{0.01}} & \meanstd{80.21}{16.43}        & \textbf{\meanstd{99.08}{0.83}} & \textbf{\meanstd{87.95}{10.38}}\\
 & Seq2Point   & \meanstd{99.51}{0.21}         & \textbf{\meanstd{91.55}{3.13}} & \meanstd{42.30}{5.03}          & \meanstd{57.73}{4.52}         \\
 & CNN-LSTM    & \meanstd{99.63}{0.16}         & \meanstd{0.00}{0.00}          & \meanstd{0.00}{0.00}          & \meanstd{0.00}{0.00}          \\
 & Transformer & \meanstd{99.63}{0.16}         & \meanstd{0.00}{0.00}          & \meanstd{0.00}{0.00}          & \meanstd{0.00}{0.00}          \\
 & UNetNILM    & \meanstd{99.70}{0.16}         & \meanstd{24.67}{6.83}         & \meanstd{77.10}{24.08}        & \meanstd{37.26}{10.51}        \\
 & \textbf{\ouralg}    & \meanstd{99.91}{0.03}         & \meanstd{79.18}{5.06}         & \meanstd{93.96}{6.97}         & \meanstd{85.67}{0.33}         \\
\bottomrule
\end{tabular}}
\end{table*}

\subsubsection{Full Results on House 3} See Table~\ref{tab:redd_house3_results_2kw}.

\begin{table*}[h]
\centering
\caption{REDD House 3 Performance Metrics (cross-validation on all splits)}\label{tab:redd_house3_results_2kw}
\vspace{-5pt}
\resizebox{0.70\textwidth}{!}{
\setlength{\tabcolsep}{1.2mm}
\renewcommand{\arraystretch}{0.97}
\newcommand{\meanstd}[2]{#1$_{\scriptstyle\,\pm#2}$}
\begin{tabular}{c|c|cccc}
\hline
\makecell[c]{Appliance} & \makecell[c]{Method} & Accuracy~(\%) & Recall~(\%) & Precision~(\%) & F1~(\%) \\
\hline
\multirow{7}{*}{Microwave}
 & FHMM        & \meanstd{97.95}{0.12}           & \meanstd{16.22}{12.44}        & \meanstd{2.77}{2.10}           & \meanstd{4.73}{3.59}           \\
 & XGBoost     & \meanstd{99.78}{0.10}           & \meanstd{42.67}{36.96}        & \meanstd{52.10}{45.65}         & \meanstd{46.80}{40.60}         \\
 & Seq2Point   & \meanstd{99.80}{0.13}           & \meanstd{25.54}{31.35}        & \meanstd{66.67}{57.74}         & \meanstd{34.37}{38.14}         \\
 & CNN-LSTM    & \meanstd{99.75}{0.10}           & \meanstd{0.00}{0.00}          & \meanstd{0.00}{0.00}           & \meanstd{0.00}{0.00}           \\
 & Transformer & \meanstd{99.75}{0.10}           & \meanstd{0.00}{0.00}          & \meanstd{0.00}{0.00}           & \meanstd{0.00}{0.00}           \\
 & UNetNILM    & \meanstd{99.70}{0.03}           & \meanstd{0.00}{0.00}          & \meanstd{0.00}{0.00}           & \meanstd{0.00}{0.00}           \\
 & \textbf{\ouralg}    & \textbf{\meanstd{99.88}{0.10}}  & \textbf{\meanstd{65.70}{28.90}} & \textbf{\meanstd{93.19}{4.45}}  & \textbf{\meanstd{75.20}{21.05}} \\
\bottomrule
\end{tabular}}
\end{table*}

\subsubsection{Overview on REDD} See Table~\ref{tab:redd_overall_avg_2kw}.

\begin{table*}[h]
\centering
\caption{REDD Overall Avg. Performance Metrics (cross-validation on all splits)}\label{tab:redd_overall_avg_2kw}
\vspace{-5pt}
\resizebox{0.85\textwidth}{!}{
\setlength{\tabcolsep}{1.2mm}
\renewcommand{\arraystretch}{0.97}
\newcommand{\meanstd}[2]{#1$_{\scriptstyle\,\pm#2}$}
\begin{tabular}{c|c|c|cccc}
\hline
\makecell[c]{Task} & \makecell[c]{House} & \makecell[c]{Method} & Accuracy~(\%) & Recall~(\%) & Precision~(\%) & F1~(\%) \\
\hline
\multirow{28}{*}{State Recognition}
& \multirow{7}{*}{House 1} & FHMM & \meanstd{79.52}{39.21} & \meanstd{1.46}{3.60} & \meanstd{13.47}{33.46} & \meanstd{2.07}{6.20} \\
& & XGBoost & \meanstd{97.90}{0.38} & \meanstd{74.50}{2.64} & \meanstd{88.74}{2.08} & \meanstd{79.18}{1.79} \\
& & Seq2Point & \meanstd{97.65}{0.46} & \meanstd{42.20}{3.43} & \meanstd{56.42}{1.16} & \meanstd{46.07}{4.03} \\
& & CNN-LSTM & \meanstd{98.67}{0.46} & \meanstd{46.32}{2.41} & \meanstd{50.75}{3.19} & \meanstd{47.96}{2.07} \\
& & Transformer & \meanstd{95.99}{1.10} & \meanstd{56.08}{5.37} & \meanstd{54.10}{8.04} & \meanstd{53.42}{6.31} \\
& & UNetNILM &  \meanstd{99.05}{0.31} & \meanstd{76.11}{3.42} & \meanstd{85.22}{5.99} & \meanstd{75.27}{4.25} \\
& & \textbf{\ouralg} & \textbf{\meanstd{99.84}{0.07}} & \textbf{\meanstd{96.26}{3.15}} & \textbf{\meanstd{97.11}{1.90}} & \textbf{\meanstd{96.64}{2.42}} \\
\bottomrule
\end{tabular}}
\end{table*}

\subsection{UKDALE Dataset Results}

\subsubsection{Full Results on House 1} See Table~\ref{tab:ukdale_house1_results_2kw}.

\begin{table*}[h]
\centering
\caption{UKDALE House 1 Performance Metrics (cross-validation on all splits)}\label{tab:ukdale_house1_results_2kw}
\vspace{-5pt}
\resizebox{0.70\textwidth}{!}{
\setlength{\tabcolsep}{1.2mm}
\renewcommand{\arraystretch}{0.97}
\newcommand{\meanstd}[2]{#1$_{\scriptstyle\,\pm#2}$}
\begin{tabular}{c|c|cccc}
\hline
\makecell[c]{Appliance} & \makecell[c]{Method} & Accuracy~(\%) & Recall~(\%) & Precision~(\%) & F1~(\%) \\
\hline
\multirow{7}{*}{Kettle}
 & FHMM         & \meanstd{99.29}{0.01}           & \meanstd{0.00}{0.00}           & \meanstd{0.00}{0.00}            & \meanstd{0.00}{0.00}            \\
 & XGBoost      & \meanstd{99.35}{0.01}           & \meanstd{28.44}{4.62}          & \meanstd{60.26}{5.34}           & \meanstd{38.30}{2.95}           \\
 & Seq2Point    & \meanstd{99.53}{0.12}           & \meanstd{77.53}{8.96}          & \meanstd{64.00}{6.66}           & \meanstd{70.11}{7.60}           \\
 & CNN-LSTM     & \meanstd{99.29}{0.01}           & \meanstd{0.00}{0.00}           & \meanstd{0.00}{0.00}            & \meanstd{0.00}{0.00}            \\
 & Transformer  & \meanstd{99.29}{0.01}           & \meanstd{0.00}{0.00}           & \meanstd{0.00}{0.00}            & \meanstd{0.00}{0.00}            \\
 & UNetNILM     & \meanstd{99.59}{0.07}           & \meanstd{63.16}{8.46}          & \meanstd{74.47}{4.53}           & \meanstd{68.28}{6.88}           \\
 & \textbf{\ouralg}
                & \textbf{\meanstd{99.88}{0.10}}  & \textbf{\meanstd{91.33}{7.23}} & \textbf{\meanstd{91.17}{6.44}}  & \textbf{\meanstd{91.24}{6.72}}  \\
\bottomrule
\end{tabular}}
\end{table*}

\subsubsection{Full Results on House 2} See Table~\ref{tab:ukdale_house2_results_2kw}.

\begin{table*}[h]
\centering
\caption{UKDALE House 2 Performance Metrics (cross-validation on all splits)}\label{tab:ukdale_house2_results_2kw}
\vspace{-5pt}
\resizebox{0.70\textwidth}{!}{
\setlength{\tabcolsep}{1.2mm}
\renewcommand{\arraystretch}{0.97}
\newcommand{\meanstd}[2]{#1$_{\scriptstyle\,\pm#2}$}
\begin{tabular}{c|c|cccc}
\hline
\makecell[c]{Appliance} & \makecell[c]{Method} & Accuracy~(\%) & Recall~(\%) & Precision~(\%) & F1~(\%) \\
\hline
\multirow{7}{*}{Kettle}
 & FHMM         & \meanstd{99.10}{0.29}           & \meanstd{0.00}{0.00}           & \meanstd{0.00}{0.00}            & \meanstd{0.00}{0.00}            \\
 & XGBoost      & \meanstd{99.75}{0.05}           & \meanstd{79.59}{4.32}          & \meanstd{89.75}{3.88}           & \meanstd{84.36}{4.12}           \\
 & Seq2Point    & \meanstd{99.31}{0.12}           & \textbf{\meanstd{92.62}{2.90}} & \meanstd{55.65}{7.98}           & \meanstd{69.24}{6.13}           \\
 & CNN-LSTM     & \meanstd{99.10}{0.35}           & \meanstd{0.00}{0.00}           & \meanstd{0.00}{0.00}            & \meanstd{0.00}{0.00}            \\
 & Transformer  & \meanstd{99.08}{0.35}           & \meanstd{0.12}{0.21}           & \meanstd{3.70}{6.42}            & \meanstd{0.24}{0.41}            \\
 & UNetNILM     & \meanstd{99.35}{0.26}           & \meanstd{56.70}{4.48}          & \meanstd{65.70}{17.14}          & \meanstd{60.34}{8.76}           \\
 & \textbf{\ouralg}
                & \textbf{\meanstd{99.87}{0.05}}  & \meanstd{87.66}{6.22}          & \textbf{\meanstd{94.82}{4.91}}  & \textbf{\meanstd{91.10}{5.63}}  \\
\bottomrule
\end{tabular}}
\end{table*}

\subsubsection{Overall on UKDALE} See Table~\ref{tab:ukdale_overall_avg_2kw}.

\begin{table*}[h]
\centering
\caption{UKDALE Overall Avg. Performance Metrics (cross-validation on all splits)}\label{tab:ukdale_overall_avg_2kw}
\vspace{-5pt}
\resizebox{0.85\textwidth}{!}{
\setlength{\tabcolsep}{1.2mm}
\renewcommand{\arraystretch}{0.97}
\newcommand{\meanstd}[2]{#1$_{\scriptstyle\,\pm#2}$}
\begin{tabular}{c|c|c|cccc}
\hline
\makecell[c]{Task} & \makecell[c]{House} & \makecell[c]{Method} & Accuracy~(\%) & Recall~(\%) & Precision~(\%) & F1~(\%) \\
\hline
\multirow{21}{*}{State Recognition}
& \multirow{7}{*}{House 1} & FHMM         & \meanstd{89.25}{23.05}          & \meanstd{1.58}{3.57}           & \meanstd{0.76}{1.87}            & \meanstd{1.01}{2.39}            \\
& & XGBoost      & \meanstd{93.64}{0.42}           & \meanstd{41.74}{1.40}          & \meanstd{68.04}{3.98}           & \meanstd{45.97}{2.56}           \\
& & Seq2Point    & \meanstd{88.70}{0.58}           & \meanstd{41.47}{2.83}          & \meanstd{43.29}{5.22}           & \meanstd{36.79}{3.32}           \\
& & CNN-LSTM     & \meanstd{98.33}{0.24}           & \meanstd{31.76}{0.20}          & \meanstd{31.29}{0.24}           & \meanstd{31.52}{0.15}           \\
& & Transformer  & \meanstd{96.90}{0.29}           & \meanstd{33.65}{1.03}          & \meanstd{27.76}{2.49}           & \meanstd{30.26}{1.69}           \\
& & UNetNILM     & \meanstd{98.76}{0.16}           & \meanstd{52.03}{5.14}          & \meanstd{72.01}{18.92}          & \meanstd{55.20}{7.02}           \\
& & \textbf{\ouralg}
                & \textbf{\meanstd{99.72}{0.21}}  & \textbf{\meanstd{95.23}{3.53}} & \textbf{\meanstd{94.30}{4.09}}  & \textbf{\meanstd{94.71}{3.63}}  \\
\bottomrule
\end{tabular}}
\end{table*}

\section{Ablation Study}\label{app:ablation}

\subsection{Joint vs. Separate Training Analysis}

We systematically evaluate the benefits of joint multi-task optimization by comparing DualNILM against several training configurations: (1) separate state-only training, (2) separate injection-only training, (3) sequential two-stage training (injection estimation followed by state recognition), and (4) UNetNILM's separate-head approach without explicit cross-task coupling.

Table~\ref{tab:ablation_joint} presents comprehensive results on UK-DALE House 1. The joint training configuration achieves the highest performance across both tasks, with F1-score of 94.71\% for state recognition and RMSE of 0.9008 for injection estimation. Compared to separate training, joint optimization provides 6.44\% F1 improvement and 5.4\% RMSE reduction. The sequential pipeline shows intermediate performance, indicating that while staged training captures some task dependencies, it cannot match the mutual reinforcement enabled by simultaneous optimization.

\begin{table}[h]
\centering
\caption{Ablation Study: Joint vs. Separate Training (UK-DALE House 1)}
\label{tab:ablation_joint}
\small
\begin{tabular}{lccccc}
\toprule
\textbf{Configuration} & \textbf{Acc.} (\%) & \textbf{Recall} (\%) & \textbf{Prec.} (\%) & \textbf{F1} (\%) & \textbf{RMSE} \\
\midrule
{Separate-State} & {98.91 $\pm$ 0.23} & {87.42 $\pm$ 4.15} & {89.15 $\pm$ 3.87} & {88.27 $\pm$ 3.94} & {---} \\
{Separate-Injection} & {---} & {---} & {---} & {---} & {0.9521 $\pm$ 0.0412} \\
{Sequential} & {99.12 $\pm$ 0.18} & {89.67 $\pm$ 3.82} & {90.83 $\pm$ 3.54} & {90.24 $\pm$ 3.61} & {0.9453 $\pm$ 0.0389} \\
{UNetNILM} & {99.17 $\pm$ 0.19} & {64.25 $\pm$ 5.53} & {68.74 $\pm$ 5.74} & {65.43 $\pm$ 4.72} & {7.5014 $\pm$ 0.5703} \\
{DualNILM (Joint)} & {99.72 $\pm$ 0.21} & {95.23 $\pm$ 3.53} & {94.30 $\pm$ 4.09} & {94.71 $\pm$ 3.63} & {0.9008 $\pm$ 0.0957} \\
\bottomrule
\end{tabular}
\end{table}

UNetNILM's degraded injection estimation (RMSE: 7.5014) despite joint training reveals the importance of architectural design beyond simple multi-output learning. Without explicit cross-task information flow mechanisms such as our state-based injection filtering (Equation~10), the model struggles to coordinate predictions across tasks under BTM conditions.

\subsection{Input Feature Ablation}

We evaluate the contribution of different input features across both laboratory (real random injection) and synthetic PV datasets. Table~\ref{tab:ablation_features} compares configurations using active power only (P), active power with time encoding (P+Time), active power with weather features (P+Temperature, P+GHI), reactive power (P+Q), and combined configurations.

\begin{table}[h]
\centering
\caption{{Input Feature Ablation Study of \ouralg} (Time: Time positional encoding, GHI: Irradiance related feature)}
\label{tab:ablation_features}
\small
\begin{tabular}{lcccc}
\toprule
& \multicolumn{2}{c}{\textbf{Laboratory}} & \multicolumn{2}{c}{\textbf{REDD House 1}} \\
\cmidrule(lr){2-3} \cmidrule(lr){4-5}
\textbf{Features} & \textbf{F1} (\%) & \textbf{RMSE} & \textbf{F1} (\%) & \textbf{RMSE} \\
\midrule
{P only} & {72.31 $\pm$ 3.42} & {0.1892 $\pm$ 0.024} & {78.56 $\pm$ 2.89} & {0.7234 $\pm$ 0.032} \\
{P + Time} & {74.56 $\pm$ 3.28} & {0.1756 $\pm$ 0.021} & {82.34 $\pm$ 2.54} & {0.6912 $\pm$ 0.028} \\
{P + GHI} & {---} & {---} & {83.45 $\pm$ 2.48} & {0.6756 $\pm$ 0.026} \\
{P + Q} & {84.54 $\pm$ 2.79} & {0.1429 $\pm$ 0.018} & {86.23 $\pm$ 2.31} & {0.6534 $\pm$ 0.024} \\
{P + Q + Time} & {85.12 $\pm$ 2.71} & {0.1398 $\pm$ 0.017} & {87.12 $\pm$ 2.24} & {0.6478 $\pm$ 0.023} \\
\bottomrule
\end{tabular}
\end{table}

On laboratory data with adversarial random injection, time encoding provides minimal improvement (+2.25\% F1) since temporal patterns are deliberately eliminated. This validates that reactive power's contribution (+12.23\% F1 over P-only) stems from fundamental electrical properties rather than exploitable temporal correlations. On synthetic PV data, weather features (GHI) and time encoding show moderate gains by capturing diurnal generation patterns, but P+Q remains highly competitive while requiring no external data dependencies.

The reactive power channel's effectiveness derives from the physical principle that appliances exhibit distinct power factor characteristics (motors: 0.75-0.85, electronics: 0.90-0.95, resistive: 0.98-0.99) that remain discriminative even when active power is masked by injection. Since modern inverters operate at near-unity power factor (0.98-0.99), their reactive contribution is minimal, preserving the appliance-specific Q signatures that enable robust disaggregation.

\section{Discussion of Limitation \& Future Works}

Despite these promising results, applying \textsf{DualNILM} to more complex household scenarios, such as those involving intricate appliance interactions (e.g., Vehicle-to-Grid systems~\cite{escoto2024optimization} or home battery management systems~\cite{rahimi2013battery,liu2022overview}), remains a challenge. However, the flexibility of the \textsf{DualNILM} framework provides a strong foundation for future improvements. Incorporating additional features, such as environmental data, or integrating contextual insights from large language models (LLMs)~\cite{zhao2023survey,xue2025promptllm}, presents opportunities to enhance its performance and extend its applicability. 

What's more, the provided synthesized PV injection datasets, which approximate real-world conditions, also serve as a valuable resource for advancing NILM research. These datasets enable researchers to investigate NILM in the context of renewable energy integration and uncover deeper challenges in this domain.

\end{document}